\documentclass{article}
\usepackage{xcolor}
\usepackage{microtype}
\usepackage{graphicx}
\usepackage{booktabs} 
\usepackage{algorithm}
\usepackage{algorithmic}
\usepackage{bbding}
\usepackage{amsmath}
\usepackage{amssymb}
\usepackage{mathtools}
\usepackage{amsthm}
\usepackage{tabularx}
\usepackage{hyperref}

\usepackage[accepted]{icml2024}

\usepackage[capitalize,noabbrev]{cleveref}

\theoremstyle{plain}
\newtheorem{theorem}{Theorem}[section]

\theoremstyle{definition}
\newtheorem{definition}[theorem]{Definition}

\theoremstyle{remark}

\usepackage[textsize=tiny]{todonotes}
\usepackage{multirow}  
\usepackage{subfig} 
\icmltitlerunning{AutoLINC}

\begin{document}

\twocolumn[
\icmltitle{Automated Loss function Search for Class-imbalanced Node Classification}




\begin{icmlauthorlist}
\icmlauthor{Xinyu Guo}{sch}
\icmlauthor{Kai Wu}{sch}
\icmlauthor{Xiaoyu Zhang}{sch2}
\icmlauthor{Jing Liu}{sch3}
\end{icmlauthorlist}

\icmlaffiliation{sch}{School of Artificial Intelligence, Xidian University, Xi'an, China}
\icmlaffiliation{sch2}{School of Cyber Engineering, Xidian University, Xi'an, China}
\icmlaffiliation{sch3}{Guangzhou Institute of Technology, Xidian University, Guangzhou, China}

\icmlcorrespondingauthor{Kai Wu}{kwu@xidian.edu.cn}

\icmlkeywords{Automated Loss Function, Monte Carlo tree search, Class-imbalanced Node Classification}

\vskip 0.3in
]



\printAffiliationsAndNotice{}  

\begin{abstract}
Class-imbalanced node classification tasks are prevalent in real-world scenarios. Due to the uneven distribution of nodes across different classes, learning high-quality node representations remains a challenging endeavor. The engineering of loss functions has shown promising potential in addressing this issue. It involves the meticulous design of loss functions, utilizing information about the quantities of nodes in different categories and the network's topology to learn unbiased node representations. However, the design of these loss functions heavily relies on human expert knowledge and exhibits limited adaptability to specific target tasks. In this paper, we introduce a high-performance, flexible, and generalizable automated loss function search framework to tackle this challenge. Across 15 combinations of graph neural networks and datasets, our framework achieves a significant improvement in performance compared to state-of-the-art methods. Additionally, we observe that homophily in graph-structured data significantly contributes to the transferability of the proposed framework.
\end{abstract}

\section{Introduction}

In recent years, the significance of learning qualitative node representations has grown in the context of accurately classifying node properties within real-world graphs \cite{wu2021a,xu2018how,zhou2020aiopen}. The adoption of graph neural networks (GNNs) \cite{kipf2017semisupervised,hamilton2017inductive,veličković2018graph} for handling graph-structured data has garnered substantial success across various domains. Nevertheless, inherent class imbalances in natural graphs can introduce a bias toward major classes, resulting in reduced accuracy for minor classes when these imbalances are not addressed.

To mitigate the challenges of class-imbalanced node classification, various methodologies have been explored \cite{ma2023class}. Notably, engineered loss functions have shown promise \cite{chen2021topology,song2022tam}, offering tailored solutions to combat class imbalance. Recent works like ReNode \cite{chen2021topology} and TAM \cite{song2022tam} have integrated graph topology information into their loss function designs. Another line of research explores the integration of contrastive learning into the context of class-imbalanced node classification \cite{zeng2023imgcl,qian2022co}. However, these approaches often rely heavily on human expert knowledge and may exhibit limited adaptability to specific target tasks.

\begin{figure}[t]
     \centering
    \includegraphics[width=\linewidth]{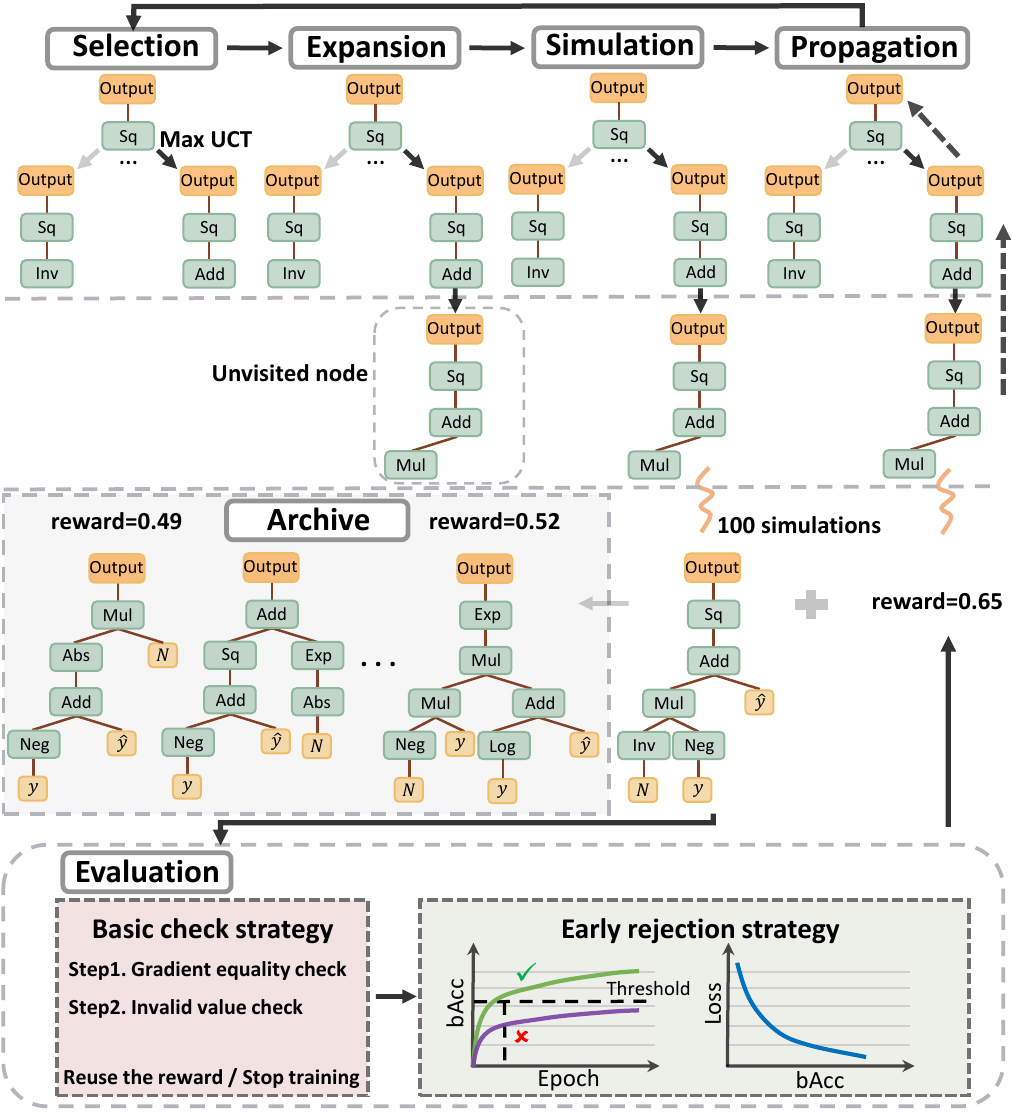}
    \caption{A schematic diagram of the AutoLINC framework with two main modules. The first module is the Monte Carlo tree search, which iteratively performs selection, expansion, simulation, and backpropagation steps to find the optimal loss function. The second module is the loss function check strategy. AutoLINC proactively filter out low-quality loss functions through the loss inspection strategy when evaluating the loss function.}
    \vskip -0.2in
    \label{fig:1}
\end{figure}

To address these challenges, this paper presents an automated framework for searching loss functions, called AutoLINC (AutoLoss for Imbalanced Node Classification). AutoLINC defines a search space uniquely suited to the task and leverages the Monte Carlo Tree Search (MCTS) algorithm to discover effective loss functions. We introduce both a Basic Check Strategy and an Early Rejection Strategy to expedite the search process. The performance of our AutoLINC framework is validated through node classification experiments involving GCN \cite{kipf2017semisupervised}, GAT \cite{veličković2018graph}, and GraphSAGE \cite{hamilton2017inductive} (abbreviated as SAGE) in three citation networks and two Amazon co-purchase networks. AutoLINC demonstrates substantial performance improvements when compared to state-of-the-art (SOTA) loss functions and non-loss function engineering methods.

The loss function search framework most closely aligned with AutoLINC's objectives is Autoloss-Zero \cite{li2021autoloss}, based on Regularized Evolution, which purports to address generic tasks. However, it faces challenges in direct application to class-imbalanced node classification problems. AutoLINC has tailored a search space, proxy tasks, and acceleration strategies specifically attuned to the complexities of imbalanced node classification problems. The comparative results against Autoloss-Zero underscore the strengths of our approach. Our contributions can be summarized as follows.
\begin{enumerate}
    \item We introduce an automated loss function search framework with the objective of transcending the limitations associated with manually crafted loss functions. AutoLINC autonomously explores loss functions tailored to class-imbalanced node classification tasks, resulting in substantial performance improvements.
    \item AutoLINC can be easily extended to other task scenarios by simply adjusting the search space and proxy tasks, making it a versatile framework. In addition to current Autoloss frameworks, we present a new high-performance loss function search framework for the Autoloss domain.
    \item Within the AutoLINC framework, we have fashioned a search space specifically designed for addressing class-imbalanced node classification issues. Additionally, we have made refinements to the MCTS algorithm to expedite the quest for optimal loss functions.
    \item We emphasize the noteworthy impact of homophily in graph-structured data on the adaptability of these loss functions to a wide range of graph datasets. Furthermore, AutoLINC demonstrates resilient generalization capabilities across varying class imbalance ratios.
\end{enumerate}

\section{Related Work}
\textbf{Class-imbalanced Node Classification}. There is currently a substantial body of work dedicated to addressing the task of semi-supervised imbalanced node classification \cite{GraphSMOTE,park2022graphens,liu2023topological,ma2023class}. Recently, the field has seen a resurgence in the application of loss function engineering, which has yielded improved classification performance.
ReNode \cite{chen2021topology}, for instance, recalibrates the impact of labeled nodes based on their proximity to class boundaries. In contrast, TAM \cite{song2022tam} adapts to the local topology of individual nodes by dynamically adjusting margins for topologically improbable instances. However, constrained by the limitations of expert knowledge, the design of loss functions is often challenging to adequately account for the characteristics of both Graph and GNNs, thereby restricting their performance.

\textbf{Loss Function Learning}. Automated loss learning aims to alleviate the considerable human effort and expertise traditionally required for loss function design. While several studies \cite{xu2018autoloss,li2019lfs,wang2020loss,liu2020stochastic,li2020auto,gao2022loss} have sought to learn loss functions automatically, they still heavily rely on human expertise in the loss search process, often initiating their search from existing loss functions. In related efforts, \cite{liu2021loss,li2021autoloss,raymond2023learning} have employed evolutionary algorithms to search for loss functions composed of primitive mathematical operators for various computer vision tasks. In the realm of loss function learning for recommendation systems, Zhao \emph{et al.} \cite{zhao2021autoloss} have introduced a framework for discovering an appropriate loss function for specific data examples, utilizing a set of base loss functions and dynamically adjusting their weights for loss combination. In contrast, Li \emph{et al.} \cite{li2022autolossgen} focus on the generation of entirely new loss functions rather than combining existing ones. However, it's worth noting that, in the context of graph data, there have been no prior attempts to apply loss function learning, necessitating the redesign of search spaces and algorithms to accommodate the unique characteristics of graph data.

\section{Preliminary}

\textit{Notation}. Given an graph $G=\{V, E\}$, where $V=\{v_1,\cdots,v_C\}$ is a set of $C$ nodes, and $E$ is a set of edges. The node feature matrix is represented as $X \in \mathbb{R}^{C\times d}$, where $d$ is the node's feature dimension. Node labels are denoted as $Y=\{y_1,\cdots,y_C|y_{i}\in\left\{1,\cdots, K\right\}\}$, where $K$ is the number of classes.

\textit{Imbalanced Node Classification}. The aim of the semi-supervised class-imbalanced node classification task is to train a classifier on an imbalanced node set $X_{train}$, typically employing a graph neural network denoted as $f_\theta$, to generate unbiased predictions $\hat{y}$ for the remaining nodes. In this context, $\theta$ symbolizes the parameters of the GNN.

\begin{definition}
\textbf{Context-free Grammar (CFG)}\citep{hopcroft2006automata,kusner2017grammar} is a formal grammar characterized by a tuple comprised of 4 elements, namely, $\mathcal{G}=(U, \Sigma, R, O)$, where $U$ denotes a finite set of non-terminal nodes, $\Sigma$ a finite set of terminal nodes, $R$ a finite set of production rules, each interpreted as a mapping from a single non-terminal symbol in $U$ to one or multiple terminal/non-terminal node(s) in $(U\cup \Sigma )^{*}$ where $*$ represents the Kleene star operation, and $O$ a single non-terminal node standing for a start symbol.
\end{definition}

\section{AutoLINC}
\subsection{Problem Definition}
Given an imbalanced graph dataset, a GNN model, and a chosen metric, AutoLINC can autonomously search for a suitable loss function $\mathcal{L}$ to train the GNN, enabling the GNN to achieve competitive performance on the test set.

\begin{definition}
The problem of AutoLoss for imbalanced node classification can be framed as a nested optimization problem, given a measure $\sigma$:
\begin{equation}
\begin{matrix}
\arg \max_\mathcal{L}  & r\left(\mathcal{L}\right)=\sigma\left(f_{\theta^\ast\left(\mathcal{L}\right)}\left(X_{val\mathrm{\ }}\right), Y_{val}\right)\\
\mathrm{\ s.t.\ }&\theta^\ast\left(\mathcal{L}\right)=\arg \min_{\theta} \mathcal{L}\left(f_\theta\left(X_{train\mathrm{\ }}\right),Y_{train}\right)\\
\end{matrix}
\label{eq:1}
\end{equation}
where $r(\mathcal{L})$ is the reward of loss function $\mathcal{L}$, $X_{val}$ is the validation dataset, $Y_{val}$ is the label of the validation dataset, and $Y_{train}$ is the label of the training dataset.
\end{definition}

\subsection{Search Space}
\label{ss}
\textbf{Input Nodes}: We take the model's output logits $\hat{y}$ and their corresponding labels $y$ as input nodes, supplemented with some constants to enhance the flexibility of the search space. Moreover, for class-imbalanced problems, we also introduce the category node count $N$ as an input node. 

\begin{table}[htbp]
\vskip -0.1in
\small
\centering
\caption{Primitive Operators.}
\vskip -0.1in
\begin{sc}
\begin{tabular}{ccc}
\toprule
Element-wise &	Expression	& Arity\\
\midrule
Add &	$x+y$	& 2\\
Mul	& $x\times y$	& 2\\
Neg	& $-x$ & 1\\
Abs	& $\left|x\right|$&	1\\
Inv&	$1/{(x+\epsilon)}$ &	1\\
Log&	$sign(x)\bullet\log(\left|x\right|+\epsilon)$ &	1\\
Exp	& $e^x$ &	1\\
Tanh&	$\tanh{x}$ &	1\\
Square&	$x^2$	&1\\
Sqrt&	$sign(x)\bullet\sqrt{\left|x\right|+\epsilon}$ &	1\\
\midrule
Aggregationr&	Expression	&Arity\\
\midrule
Mean &	$\frac{1}{C}\sum_{i=1}^{C}\left(x_i\right)$ &	1\\
\bottomrule
\end{tabular}
\end{sc}
\vskip -0.1in
\label{tab:io}
\end{table}

\textbf{Primitive Operators}: 
The primitive operators consist of element-wise operators and aggregation operator, as shown in Appendix Table \ref{tab:io}. 
We use only one aggregation operator, Mean, as the final aggregation. 

\textbf{Solution Expression}: 
Loss functions are fundamentally mathematical expressions that can be represented in the form of parse trees using a CFG. In our work, using the elements we have, $U$ denotes symbols like ($D$). $O$ corresponds to the output of the loss function (Output, $o$). $\Sigma$ is $\{y, \hat{y}, N\}$. $R$ is our primitive operators, mapping from one non-terminal node (e.g., $o$/$D$) to another node. The parse tree of each loss function starts from the root node, Output, and traverses pre-order based on the product rule until all leaf nodes are represented by symbols from the terminal node set. As shown in Fig. \ref{fig:1}, the loss function is presented in the form of a parse tree (for clarity, non-terminal node $D$ is hidden, while the product rules are retained). When an expression is evaluated, the model's output $\hat{y}$ has the same shape as the one-hot encoded labels $y$ and the number of categories $N$. The tree's output is eventually averaged to obtain a single value after the leaf nodes input into the loss function expression tree are calculated. 

\subsection{Search Algorithm}
\label{sa}

Unlike the current Autoloss search framework based on Regularized evolution \cite{real20a,real2019regularized}, we design a more capable search algorithm. AutoLINC primarily consists of two parts: 1) MCTS \cite{coulom2006efficient,sun2023symbolic}, which explores the most promising expressions through repeated selection, expansion, simulation, and backpropagation steps; 2) Task evaluation with a loss check strategy, which filters out evaluated, training-unfriendly, poorly converging, and poor-performing loss functions, speeding up the evaluation of proxy tasks. AutoLINC's framework is illustrated in Fig. \ref{fig:1}, and the detailed sections can be found in Algorithm 1. MCTS repeats the below steps until the termination conditions are met.

\begin{table*}[htbp]
  \centering
  \caption{Compared results of AutoLINC with the SOTA loss functions. Values in bold indicate the best performance, while values underlined indicate the second-best performance. $*\pm*$ denotes the mean and standard errors.}
  \begin{sc}
  \resizebox{\linewidth}{!}{
    \begin{tabular}{cl|ll|ll|ll|ll|ll}
\toprule
\multirow{2}[2]{*}{GNN} & \textbf{Dataset} & \multicolumn{2}{c|}{Cora} & \multicolumn{2}{c|}{CiteSeer} & \multicolumn{2}{c|}{PubMed} & \multicolumn{2}{c|}{Amazon-computers} & \multicolumn{2}{c}{Amazon-photo} \\
          & \textbf{Imbalance Ratio:10} & \multicolumn{1}{c}{bAcc} & \multicolumn{1}{c|}{F1} & \multicolumn{1}{c}{bAcc} & \multicolumn{1}{c|}{F1} & \multicolumn{1}{c}{bAcc} & \multicolumn{1}{c|}{F1} & \multicolumn{1}{c}{bAcc} & \multicolumn{1}{c|}{F1} & \multicolumn{1}{c}{bAcc} & \multicolumn{1}{c}{F1} \\
    \midrule
    \multirow{7}[4]{*}{GCN} & CE    & 53.89$\pm$0.77 & 49.13$\pm$1.20 & 35.93$\pm$0.45 & 23.58$\pm$0.74 & 61.96$\pm$0.51 & 47.55$\pm$1.01 & 80.99$\pm$0.69 & 72.67$\pm$0.48 & 78.77$\pm$1.08 & 73.17$\pm$1.53 \\
          & Re-Weight & 60.91$\pm$1.05 & 59.18$\pm$1.31 & 40.79$\pm$1.56 & 31.95$\pm$1.95 & 67.18$\pm$1.20 & 59.74$\pm$2.19 & 82.82$\pm$0.36 & 73.44$\pm$0.43 & 81.87$\pm$0.73 & 75.04$\pm$1.09 \\
          & PC Softmax & 68.15$\pm$0.82 & 67.90$\pm$0.91 & 49.70$\pm$1.28 & 45.24$\pm$1.33 & 70.44$\pm$0.62 & 69.21$\pm$0.76 & \textbf{85.10$\pm$0.06} & \textbf{75.20$\pm$0.17} & 78.20$\pm$0.66 & 75.41$\pm$0.66 \\
          & BalancedSoftmax & 68.96$\pm$0.52 & 68.67$\pm$0.49 & 53.57$\pm$1.44 & 50.55$\pm$1.76 & 71.00$\pm$1.12 & 69.72$\pm$1.18 & \underline{83.70$\pm$0.31} & \underline{74.48$\pm$0.48} & 81.93$\pm$0.95 & 76.19$\pm$1.55 \\
          & BalancedSoftmax+TAM & \underline{69.17$\pm$0.77} & \underline{69.00$\pm$0.73} & \underline{55.65$\pm$1.19} & \underline{54.06$\pm$1.39} & \underline{72.15$\pm$1.08} & \underline{71.79$\pm$1.35} & 83.47$\pm$0.32 & 74.40$\pm$0.50 & \underline{82.50$\pm$0.79} & \underline{76.25$\pm$1.41} \\
          & ReNode & 64.64$\pm$1.09 & 64.00$\pm$1.43 & 42.02$\pm$1.39 & 33.36$\pm$1.82 & 68.58$\pm$1.19 & 62.97$\pm$2.51 & 82.75$\pm$0.22 & 72.96$\pm$0.37 & 81.68$\pm$0.75 & 74.82$\pm$1.01 \\
\cmidrule{2-12}          & AutoLINC & \textbf{70.21$\pm$0.67} & \textbf{69.67$\pm$0.79} & \textbf{56.56$\pm$1.97} & \textbf{56.09$\pm$2.06} & \textbf{72.43$\pm$1.03} & \textbf{72.42$\pm$1.07} & 83.23$\pm$0.23 & 71.83$\pm$0.45 & \textbf{83.06$\pm$0.80} & \textbf{76.66$\pm$1.18} \\
    \midrule
    \multirow{7}[4]{*}{GAT} & CE    & 49.26$\pm$0.97 & 44.80$\pm$1.36 & 34.34$\pm$0.42 & 21.86$\pm$0.35 & 60.31$\pm$0.23 & 46.41$\pm$0.52 & 66.68$\pm$0.88 & 61.28$\pm$0.88 & 57.36$\pm$1.15 & 51.26$\pm$1.03 \\
          & Re-Weight & \underline{67.27$\pm$0.67} & \underline{66.35$\pm$0.72} & 44.73$\pm$1.64 & 39.73$\pm$2.19 & 67.89$\pm$1.18 & 60.83$\pm$2.15 & 73.66$\pm$0.65 & 66.61$\pm$0.98 & 64.06$\pm$1.16 & 58.94$\pm$1.25 \\
          & PC Softmax & 64.67$\pm$1.02 & 63.60$\pm$1.22 & 50.50$\pm$1.59 & 45.86$\pm$1.94 & \underline{72.49$\pm$0.62} & 70.97$\pm$0.77 & 72.18$\pm$1.16 & 64.80$\pm$1.15 & \underline{66.49$\pm$1.39} & \underline{62.57$\pm$1.75} \\
          & BalancedSoftmax & 65.92$\pm$0.70 & 65.43$\pm$0.82 & 53.70$\pm$1.56 & 50.31$\pm$1.93 & 71.24$\pm$0.84 & 69.66$\pm$0.91 & 74.88$\pm$0.53 & 68.84$\pm$0.86 & 64.81$\pm$1.05 & 60.09$\pm$1.35 \\
          & BalancedSoftmax+TAM & 66.30$\pm$1.01 & 65.28$\pm$1.03 & \underline{54.14$\pm$1.31} & \underline{51.84$\pm$1.63} & 72.24$\pm$0.85 & \underline{71.96$\pm$0.76} & \underline{75.66$\pm$0.62} & \underline{69.80$\pm$1.09} & 65.88$\pm$0.98 & 61.47$\pm$0.99 \\
          & ReNode & 64.19$\pm$1.20 & 64.01$\pm$1.42 & 41.15$\pm$1.26 & 35.45$\pm$1.49 & 68.19$\pm$1.46 & 62.78$\pm$2.71 & 75.17$\pm$0.79 & 67.47$\pm$1.04 & 63.54$\pm$1.43 & 59.27$\pm$1.58 \\
\cmidrule{2-12}          & AutoLINC & \textbf{70.63$\pm$1.01} & \textbf{70.16$\pm$1.04} & \textbf{58.48$\pm$1.57} & \textbf{57.38$\pm$1.89} & \textbf{72.57$\pm$0.88} & \textbf{72.27$\pm$1.02} & \textbf{78.86$\pm$0.59} & \textbf{71.73$\pm$0.73} & \textbf{74.41$\pm$0.69} & \textbf{69.11$\pm$0.91} \\
    \midrule
    \multirow{7}[3]{*}{SAGE} & CE    & 51.57$\pm$0.54 & 45.11$\pm$0.83 & 35.34$\pm$0.25 & 22.82$\pm$0.38 & 61.01$\pm$0.55 & 48.03$\pm$0.88 & 68.51$\pm$1.52 & 62.78$\pm$1.78 & 59.74$\pm$1.67 & 51.53$\pm$1.96 \\
          & Re-Weight & 55.17$\pm$0.88 & 52.07$\pm$1.33 & 38.53$\pm$1.35 & 30.60$\pm$1.91 & 62.00$\pm$0.87 & 52.05$\pm$2.03 & 72.41$\pm$1.05 & 63.77$\pm$0.87 & 60.28$\pm$1.47 & 52.03$\pm$1.74 \\
          & PC Softmax & 65.49$\pm$0.65 & 64.73$\pm$0.82 & 49.54$\pm$1.45 & 45.04$\pm$1.82 & \textbf{71.38$\pm$0.77} & \textbf{70.52$\pm$1.04} & 75.17$\pm$0.54 & 68.78$\pm$0.48 & 64.04$\pm$0.74 & 58.62$\pm$1.09 \\
          & BalancedSoftmax & 65.38$\pm$0.66 & 65.06$\pm$0.76 & 51.56$\pm$1.68 & 48.60$\pm$1.91 & 69.77$\pm$0.70 & 68.63$\pm$0.83 & 75.76$\pm$1.02 & 68.03$\pm$0.74 & 67.62$\pm$1.86 & 60.96$\pm$2.09 \\
          & BalancedSoftmax+TAM & \underline{66.54$\pm$0.49} & \underline{66.24$\pm$0.65} & \underline{52.46$\pm$1.23} & \underline{49.27$\pm$1.72} & 70.26$\pm$0.87 & 69.94$\pm$2.83 & \underline{78.11$\pm$0.77} & \underline{70.53$\pm$0.65} & \underline{69.45$\pm$2.01} & \underline{62.64$\pm$2.17} \\
          & ReNode & 59.36$\pm$0.57 & 57.70$\pm$0.48 & 41.12$\pm$1.56 & 32.61$\pm$1.80 & 64.67$\pm$1.21 & 55.82$\pm$2.72 & 76.25$\pm$0.71 & 69.03$\pm$0.56 & 64.72$\pm$1.25 & 57.16$\pm$1.35 \\
\cmidrule{2-12}          & AutoLINC & \textbf{68.51$\pm$1.02} & \textbf{68.49$\pm$1.02} & \textbf{56.78$\pm$1.27} & \textbf{56.16$\pm$1.37} & \underline{70.97$\pm$1.04} & \underline{70.32$\pm$0.96} & \textbf{80.02$\pm$0.68} & \textbf{71.19$\pm$0.32} & \textbf{72.98$\pm$1.53} & \textbf{66.66$\pm$1.99} \\
\bottomrule
    \end{tabular}}
    \end{sc}
  \label{tab:1}
\end{table*}

\textbf{Step 1: Selection} - MCTS starts from the root node and uses the Upper Confidence Bounds applied for Trees (UCT) selection strategy to choose the next node, repeatedly selecting until reaching a leaf node or an expandable node.
\begin{equation}
\label{eq:UCT}
    UCT(s,a) = Q(s,a) + c \sqrt{\ln[W(s)]/W(s,a)}
\end{equation}
Here, $Q(s,a)$ represents the average reward of taking action $a \in \mathcal{A}$ in state $s$. $Q(s,a)$ encourages exploitation of the current best child node. In this paper, our objective is to determine the optimal expression for the loss function. To achieve this, we define the maximum simulated result value as $Q(s,a)$ in state $s$ when taking action $a$. $W(s)$ is the number of visits to node $s$, and $W(s,a)$ is the number of times action $a$ has been taken in state $s$. Therefore, $\sqrt{\ln[W(s)]/W(s,a)}$ encourages exploration of other nodes. $c$ is the exploration rate, generally defined empirically for a specific problem. In the selection, expansion, or simulation process, we limit the maximum number for selecting product rules from the root node to 10 to avoid lengthy loss function expressions.

\textbf{Step 2: Expansion} - Upon reaching an expandable node, a randomly unvisited child node is chosen for expansion. After expansion, an evaluation is conducted if a terminal state is reached. The evaluation results are backtracked to update the parent node until the root node.

\textbf{Step 3: Simulation} - If the currently expanded node is still non-terminal, a random selection is made for simulation, continuously choosing child nodes until a terminal state is reached. In calculating rewards for loss functions reaching a terminal state, we use the bAcc calculated on the validation set as the reward.

\textbf{Step 4: Backpropagation}- Update the visit count and $Q$ value of the parent node until the root node based on the simulation results.

\subsection{Loss Function Check}

\textbf{Legality Check of the Loss Function:} A legitimate loss function should encompass the GNN's output logits $\hat{y}$ and the corresponding labels $y$. To ensure the loss function's competence in addressing class-imbalanced node classification issues, it is imperative that the function accounts for the class-specific node counts, represented as $N$. Any loss functions lacking GNN output logits $\hat{y}$, labels $y$, or class-specific node counts $N$ will not undergo evaluation.

\textbf{Basic Check Strategy:} This strategy encompasses fundamental checks, such as detecting invalid Nan values and ensuring gradient equality. Because loss functions are represented in the form of expression trees, they may exhibit symmetrical cases, where different-looking loss function trees are equivalent. Additionally, distinct loss functions can sometimes yield equivalent gradients. We maintain records of the evaluated loss function formulas and their respective rewards. In the case of equivalent formulas, their rewards are directly reused.

\textbf{Early Rejection Strategy:} While legality checks ensure that the evaluated solutions incorporate ${y, \hat{y}, N}$, and equality checks prevent the redundant evaluation of equivalent formulas, there remain a significant number of underperforming, training-unfriendly loss functions. To address this, we employ the Early Rejection Strategy to discard poorly converging and ineffective loss functions. The Early Rejection Strategy comprises Monotonicity Checks and Poor-performance Rejections.

1) \textit{Monotonicity  Check}: As per \cite{liu2021loss}, a crucial criterion during the proxy task's training is assessing the loss function's quality based on its impact on accuracy metrics. An ideal loss function should exhibit a monotonous increase in accuracy metrics on the training set as the loss value decreases. If the loss function's reduction is not positively correlated with improved accuracy metrics on the training set, it is deemed invalid. In such cases, the proxy task is prematurely terminated to mitigate computational costs.

2) \textit{Poor-performance Rejection}: Throughout the proxy task's training phase, the loss function is considered suboptimal if the metrics on the validation set significantly lag behind the performance achieved by the Top loss functions searched thus far. In such scenarios, the proxy task is concluded early to reduce computational costs.

\begin{table*}[htbp]
  \centering
  \caption{The comparison results of the loss functions discovered by AutoLINC in Table \ref{tab:exp} on datasets with $\rho=5$.}
  \begin{sc}
  \resizebox{\linewidth}{!}{
    \begin{tabular}{cl|ll|ll|ll|ll|ll}
    \toprule
\multirow{2}[2]{*}{GNN} & \textbf{Dataset} & \multicolumn{2}{c|}{Cora} & \multicolumn{2}{c|}{CiteSeer} & \multicolumn{2}{c|}{PubMed} & \multicolumn{2}{c|}{Amazon-computers} & \multicolumn{2}{c}{Amazon-photo} \\
          & \textbf{Imbalance Ratio:5} & \multicolumn{1}{c}{bAcc} & \multicolumn{1}{c|}{F1} & \multicolumn{1}{c}{bAcc} & \multicolumn{1}{c|}{F1} & \multicolumn{1}{c}{bAcc} & \multicolumn{1}{c|}{F1} & \multicolumn{1}{c}{bAcc} & \multicolumn{1}{c|}{F1} & \multicolumn{1}{c}{bAcc} & \multicolumn{1}{c}{F1} \\
    \midrule
    \multirow{7}[3]{*}{GCN} & CE    & 64.19$\pm$0.80 & 64.08$\pm$1.08 & 41.23$\pm$0.98 & 32.82$\pm$1.41 & 65.32$\pm$0.84 & 55.58$\pm$1.77 & 83.80$\pm$0.40 & 76.22$\pm$0.37 & 86.50$\pm$0.76 & 82.72$\pm$1.15 \\
          & Re-Weight & 69.74$\pm$0.98 & 70.09$\pm$1.00 & 49.49$\pm$1.62 & 45.45$\pm$2.00 & 71.52$\pm$0.91 & 67.24$\pm$1.35 & 84.34$\pm$0.30 & \underline{77.52$\pm$0.34} & 87.02$\pm$0.69 & 82.64$\pm$1.06 \\
          & PC Softmax & 73.36$\pm$0.55 & 73.35$\pm$0.57 & 59.18$\pm$1.29 & 57.85$\pm$1.45 & 73.77$\pm$0.64 & 72.55$\pm$0.68 & \textbf{85.49$\pm$0.11} & \textbf{78.73$\pm$0.23} & \textbf{87.77$\pm$0.10} & \textbf{83.99$\pm$0.10} \\
          & BalancedSoftmax & 73.98$\pm$0.81 & 73.61$\pm$0.77 & \underline{60.66$\pm$1.08} & 59.72$\pm$1.13 & \textbf{74.67$\pm$0.87} & 73.35$\pm$1.03 & \underline{84.99$\pm$0.39} & 77.48$\pm$0.29 & 87.29$\pm$0.66 & 82.92$\pm$0.91 \\
          & BalancedSoftmax+TAM & \underline{74.02$\pm$0.70} & \underline{74.08$\pm$0.67} & \textbf{61.14$\pm$1.29} & \textbf{60.41$\pm$1.31} & \underline{74.55$\pm$0.82} & \underline{73.92$\pm$0.99} & 84.88$\pm$0.38 & 77.21$\pm$0.22 & \underline{87.52$\pm$0.61} & \underline{83.21$\pm$0.88} \\
          & ReNode & 72.50$\pm$0.81 & 73.11$\pm$0.76 & 53.48$\pm$1.55 & 50.81$\pm$1.75 & 71.86$\pm$0.82 & 67.96$\pm$1.17 & 84.89$\pm$0.18 & 76.98$\pm$0.22 & 87.37$\pm$0.71 & 82.74$\pm$1.11 \\
\cmidrule{2-12}          & AutoLINC & \textbf{75.26$\pm$0.54} & \textbf{74.79$\pm$0.47} & 60.64$\pm$1.05 & \underline{60.39$\pm$1.01} & 74.42$\pm$0.86 & \textbf{74.61$\pm$0.88} & 84.63$\pm$0.37 & 73.80$\pm$0.58 & 87.03$\pm$0.83 & 81.57$\pm$1.28 \\
\midrule
    \multirow{7}[2]{*}{GAT} & CE    & 57.12$\pm$1.15 & 56.32$\pm$1.38 & 37.76$\pm$1.29 & 28.27$\pm$1.95 & 64.11$\pm$0.84 & 55.76$\pm$1.66 & 73.30$\pm$0.81 & 67.85$\pm$1.04 & 72.05$\pm$1.62 & 67.48$\pm$1.60 \\
          & Re-Weight & 72.99$\pm$0.88 & 72.05$\pm$1.08 & 53.72$\pm$1.24 & 51.61$\pm$1.52 & 72.65$\pm$0.77 & 68.80$\pm$1.14 & 76.98$\pm$0.58 & 69.89$\pm$0.78 & 77.68$\pm$0.77 & 73.72$\pm$1.21 \\
          & PC Softmax & 72.11$\pm$0.57 & 71.85$\pm$0.69 & 57.86$\pm$1.46 & 55.91$\pm$1.71 & 73.61$\pm$0.70 & 72.37$\pm$0.88 & 75.11$\pm$0.75 & 68.31$\pm$1.11 & \underline{79.85$\pm$0.90} & \textbf{75.93$\pm$0.80} \\
          & BalancedSoftmax & 73.04$\pm$0.37 & 72.44$\pm$0.43 & 58.97$\pm$1.27 & 58.20$\pm$1.33 & 74.25$\pm$0.63 & 72.82$\pm$0.77 & 78.21$\pm$0.66 & 71.52$\pm$0.74 & 75.72$\pm$0.88 & 71.54$\pm$1.21 \\
          & BalancedSoftmax+TAM & 73.37$\pm$0.67 & 72.47$\pm$0.78 & \underline{60.26$\pm$0.84} & \underline{59.48$\pm$0.87} & \textbf{74.75$\pm$0.66} & \underline{74.09$\pm$0.83} & \underline{79.04$\pm$0.89} & \underline{72.56$\pm$1.11} & 77.52$\pm$1.11 & 73.48$\pm$1.48 \\
          & ReNode & \underline{74.12$\pm$0.75} & \underline{74.44$\pm$0.77} & 54.43$\pm$1.95 & 52.82$\pm$2.41 & 72.35$\pm$1.05 & 68.89$\pm$1.76 & 78.84$\pm$0.67 & 71.56$\pm$1.01 & 76.08$\pm$1.69 & 70.89$\pm$1.91 \\
\cmidrule{2-12}          & AutoLINC & \textbf{75.57$\pm$0.42} & \textbf{75.24$\pm$0.40} & \textbf{63.58$\pm$0.76} & \textbf{63.29$\pm$0.71} & \underline{74.51$\pm$0.82} & \textbf{74.50$\pm$0.79} & \textbf{82.60$\pm$0.68} & \textbf{74.39$\pm$1.00} & \textbf{79.98$\pm$1.02} & \underline{74.39$\pm$1.75} \\
\midrule
    \multirow{7}[3]{*}{SAGE} & CE    & 58.76$\pm$0.85 & 56.39$\pm$1.22 & 39.18$\pm$0.97 & 30.11$\pm$1.56 & 63.71$\pm$0.89 & 53.82$\pm$1.70 & 80.13$\pm$0.55 & 72.55$\pm$0.82 & 78.11$\pm$1.01 & 72.51$\pm$1.50 \\
          & Re-Weight & 65.84$\pm$1.13 & 65.23$\pm$1.35 & 45.60$\pm$1.46 & 40.75$\pm$1.95 & 67.81$\pm$0.79 & 62.56$\pm$1.50 & 81.03$\pm$0.54 & 74.40$\pm$0.45 & 79.09$\pm$1.14 & 72.93$\pm$1.71 \\
          & PC Softmax & 71.20$\pm$0.96 & 71.07$\pm$1.00 & 56.82$\pm$1.63 & 55.18$\pm$1.82 & \textbf{73.69$\pm$0.32} & \textbf{72.87$\pm$0.41} & 82.42$\pm$0.36 & 72.27$\pm$0.45 & 81.77$\pm$0.29 & \underline{77.74$\pm$0.27} \\
          & BalancedSoftmax & 71.05$\pm$0.75 & 71.00$\pm$0.91 & 57.88$\pm$1.25 & 57.00$\pm$1.26 & 73.19$\pm$0.30 & 72.05$\pm$0.43 & 81.35$\pm$0.49 & \textbf{74.76$\pm$0.48} & \textbf{82.94$\pm$1.00} & \textbf{77.93$\pm$1.52} \\
          & BalancedSoftmax+TAM & \underline{71.71$\pm$0.58} & \underline{71.55$\pm$0.66} & \underline{59.55$\pm$1.29} & \underline{58.68$\pm$1.27} & \underline{73.36$\pm$0.55} & \underline{72.46$\pm$0.66} & \underline{82.46$\pm$0.32} & \underline{74.56$\pm$0.34} & 82.74$\pm$1.17 & 77.18$\pm$1.68 \\
          & ReNode & 69.30$\pm$1.40 & 69.21$\pm$1.44 & 53.14$\pm$1.53 & 51.05$\pm$1.87 & 68.39$\pm$1.09 & 62.26$\pm$2.03 & 82.41$\pm$0.32 & 73.82$\pm$0.71 & \underline{82.79$\pm$0.69} & 76.90$\pm$1.17 \\
\cmidrule{2-12}          & AutoLINC & \textbf{74.15$\pm$0.69} & \textbf{73.80$\pm$0.70} & \textbf{61.41$\pm$0.85} & \textbf{61.13$\pm$0.78} & 72.59$\pm$0.71 & 72.08$\pm$0.74 & \textbf{82.79$\pm$0.36} & 72.96$\pm$0.64 & 82.31$\pm$0.95 & 77.67$\pm$1.36 \\
\bottomrule   
\end{tabular}
}%
\end{sc}
  \label{tab:2}%
\end{table*}

\subsection{Framework of AutoLINC} To manage non-terminal nodes within the parsing tree, we employ a last-in-first-out strategy. The last non-terminal node placed on the stack, denoted as $NT$, represents the current node. We define the action space $\mathcal{A}$ as $R$ and the state space $\mathcal{S}$ as all possible traversals of complete or incomplete parse trees in ordered sequences. In the current state $s_t=[a_1, a_2, \cdots, a_t]$, the MCTS agent filters out invalid production rules for the present non-terminal node. Subsequently, it selects a valid rule as action $a_{t+1}$. This leads to the expansion of the parse tree with a new terminal or non-terminal branch, determined by $a_{t+1}$. Concurrently, the agent progresses to a new state $s_{t+1}=[a_1, a_2, \cdots, a_t, a_{t+1}]$. The agent proceeds by removing the current non-terminal symbol from $NT$ and adding any non-terminal nodes, if they exist, on the right-hand side of the selected rule to the stack. 

Additionally, legality check, basic check strategies, and early rejection are executed, followed by reward calculation based on the proxy task (Eq. \ref{eq:1}). The discovered $\mathcal{L}$ and its associated reward are recorded in $M$. When the agent encounters an unvisited node, a series of simulations commence, with the agent randomly selecting the next node until the parse tree is completed. In a similar manner, the reward is computed after performing the loss function check strategy. The highest result from these attempts serves as the reward for the current simulation phase, backpropagating from the current unvisited node to the root node. We then select the top 10 loss functions, denoted as $M_{top}$, from $M$ and extract the best loss function, $\mathcal{L}^\star$, from $M_{top}$. Note that the reward is calculated based on the complete task. Appendix Algorithm 1 describes the detailed procedure of AutoLINC.

\subsection{Proxy Task}
To accelerate the search process, rewarding loss function expressions typically necessitate training graph neural networks, a time-intensive endeavor. In evaluating these loss functions, we employ a lightweight proxy training task. Given the inherent challenges in partitioning graph data structures, we curtail the number of training iterations on real datasets for this proxy task. Subsequently, rewards are computed based on the validation set's balanced accuracy.

\section{Experiments}
We illustrate the effectiveness and adaptability of AutoLINC by addressing the following key questions:

\begin{table*}[htbp]
  \centering
  \caption{The performance of loss functions A, E, I, J, and K on various combinations of datasets and graph neural networks.}
  \begin{sc}
  \resizebox{\linewidth}{!}{
    \begin{tabular}{cl|ll|ll|ll|ll|ll}
    \toprule
    \multirow{2}[2]{*}{GNN} & \textbf{Dataset} & \multicolumn{2}{c|}{Cora} & \multicolumn{2}{c|}{CiteSeer} & \multicolumn{2}{c|}{PubMed} & \multicolumn{2}{c|}{Amazon-computers} & \multicolumn{2}{c}{Amazon-photo} \\
          & \textbf{Imbalance Ratio: 10} & \multicolumn{1}{c}{bAcc} & \multicolumn{1}{c|}{F1} & \multicolumn{1}{c}{bAcc} & \multicolumn{1}{c|}{F1} & \multicolumn{1}{c}{bAcc} & \multicolumn{1}{c|}{F1} & \multicolumn{1}{c}{bAcc} & \multicolumn{1}{c|}{F1} & \multicolumn{1}{c}{bAcc} & \multicolumn{1}{c}{F1} \\
    \midrule
    \multirow{6}[3]{*}{GCN} 
          & BalancedSoftmax+TAM & 69.17$\pm$0.77 & 69.00$\pm$0.73 & 55.65$\pm$1.19 & 54.06$\pm$1.39 & \underline{72.15$\pm$1.08} & \underline{71.79$\pm$1.35} & \textbf{83.47$\pm$0.32} & \textbf{74.40$\pm$0.50} & \underline{82.50$\pm$0.79} & \multicolumn{1}{l}{\underline{76.25$\pm$1.41}} \\
\cmidrule{2-12}          & Cora-GCN(A) & \underline{70.21$\pm$0.67} & \underline{69.67$\pm$0.79} & \underline{56.64$\pm$2.04} & \underline{55.82$\pm$2.03} & 71.95$\pm$1.12 & 71.64$\pm$1.19 & 26.82$\pm$3.29 & 17.67$\pm$3.12 & 33.40$\pm$3.84 & \multicolumn{1}{l}{24.00$\pm$4.20} \\
          & CiteSeer-GAT(E) & 69.86$\pm$0.83 & 68.77$\pm$0.74 & 56.15$\pm$2.00 & 55.56$\pm$2.02 & 71.40$\pm$1.49 & 71.17$\pm$1.48 & 35.42$\pm$5.00 & 21.68$\pm$4.72 & 35.79$\pm$4.12 & \multicolumn{1}{l}{23.91$\pm$4.56} \\
          & PubMed-SAGE(I) & \textbf{71.13$\pm$0.82} & \textbf{70.47$\pm$0.88} & \textbf{58.32$\pm$1.56} & \textbf{57.64$\pm$1.61} & \textbf{72.66$\pm$1.01} & \textbf{72.50$\pm$1.18} & 16.74$\pm$1.20 & 6.48$\pm$1.21 & 27.24$\pm$2.20 & \multicolumn{1}{l}{19.37$\pm$2.94} \\
\cmidrule{2-12}          & computers-GCN(J) & 68.41$\pm$0.70 & 67.28$\pm$0.66 & 53.05$\pm$1.36 & 51.06$\pm$1.48 & 68.03$\pm$1.02 & 61.65$\pm$1.89 & \underline{83.23$\pm$0.23} & \underline{71.83$\pm$0.45} & 80.64$\pm$0.69 & \multicolumn{1}{l}{73.80$\pm$1.01} \\
          & photo-GCN(K) & 64.50$\pm$0.34 & 63.46$\pm$0.34 & 52.93$\pm$0.76 & 51.82$\pm$0.86 & 67.18$\pm$0.80 & 60.28$\pm$1.57 & 82.36$\pm$0.21 & 70.55$\pm$0.38 & \textbf{83.06$\pm$0.80} & \multicolumn{1}{l}{\textbf{76.66$\pm$1.18}} \\
    \midrule
    \multirow{6}[3]{*}{GAT}   & BalancedSoftmax+TAM & 66.30$\pm$1.01 & 65.28$\pm$1.03 & 54.14$\pm$1.31 & 51.84$\pm$1.63 & 72.24$\pm$0.85 & \underline{71.96$\pm$0.76} & \underline{75.66$\pm$0.62} & \textbf{69.80$\pm$1.09} & \underline{65.88$\pm$0.98} & \multicolumn{1}{l}{\underline{61.47$\pm$0.99}} \\
\cmidrule{2-12}          & Cora-GCN(A) & 69.83$\pm$0.96 & 70.05$\pm$0.95 & 58.30$\pm$1.39 & 57.17$\pm$1.74 & \underline{72.80$\pm$1.06} & 71.92$\pm$1.29 & 50.85$\pm$7.11 & 43.45$\pm$6.16 & 59.37$\pm$6.79 & \multicolumn{1}{l}{54.25$\pm$6.74} \\
          & CiteSeer-GAT(E) & \underline{70.79$\pm$0.78} & \underline{70.63$\pm$0.83} & \underline{58.48$\pm$1.57} & \underline{57.38$\pm$1.89} & 72.50$\pm$1.04 & 71.55$\pm$1.03 & 40.61$\pm$7.01 & 35.02$\pm$6.54 & 55.09$\pm$7.59 & \multicolumn{1}{l}{49.64$\pm$7.84} \\
          & PubMed-SAGE(I) & \textbf{71.51$\pm$0.69} & \textbf{71.32$\pm$0.68} & \textbf{59.11$\pm$1.28} & \textbf{58.02$\pm$1.43} & \textbf{73.62$\pm$0.83} & \textbf{72.98$\pm$1.10} & 20.47$\pm$2.01 & 14.45$\pm$2.06 & 20.62$\pm$2.16 & \multicolumn{1}{l}{16.29$\pm$2.82} \\
\cmidrule{2-12}          & computers-GCN(J) & 57.82$\pm$1.24 & 56.47$\pm$1.24 & 44.01$\pm$1.27 & 41.47$\pm$1.53 & 64.25$\pm$1.04 & 56.20$\pm$1.83 & 75.15$\pm$0.73 & 67.50$\pm$0.79 & 59.90$\pm$1.66 & \multicolumn{1}{l}{55.01$\pm$1.82} \\
          & photo-GCN(K) & 60.46$\pm$0.89 & 59.04$\pm$0.85 & 45.45$\pm$1.35 & 43.07$\pm$1.59 & 66.78$\pm$0.98 & 61.24$\pm$1.46 & \textbf{77.95$\pm$0.86} & \underline{69.21$\pm$1.04} & \textbf{71.17$\pm$1.59} & \multicolumn{1}{l}{\textbf{67.49$\pm$1.42}} \\
    \midrule
    \multirow{6}[3]{*}{SAGE} & BalancedSoftmax+TAM & 66.54$\pm$0.49 & 66.24$\pm$0.65 & 52.46$\pm$1.23 & 49.27$\pm$1.72 & 70.26$\pm$0.87 & \underline{69.94$\pm$2.83} & \underline{78.11$\pm$0.77} & \underline{70.53$\pm$0.65} & \underline{69.45$\pm$2.01} & \multicolumn{1}{l}{\underline{62.64$\pm$2.17}} \\
\cmidrule{2-12}          & Cora-GCN(A) & \textbf{68.46$\pm$0.94} & \textbf{68.20$\pm$0.89} & \underline{56.68$\pm$1.24} & \underline{55.92$\pm$1.44} & 70.12$\pm$0.98 & 69.34$\pm$1.23 & 14.83$\pm$0.73 & 5.68$\pm$0.73 & 21.37$\pm$1.57 & \multicolumn{1}{l}{11.95$\pm$1.37} \\
          & CiteSeer-GAT(E) & \underline{68.21$\pm$0.98} & \underline{68.03$\pm$1.01} & \textbf{56.88$\pm$1.29} & \textbf{56.04$\pm$1.33} & \underline{70.38$\pm$0.94} & 69.00$\pm$1.16 & 14.71$\pm$1.10 & 5.33$\pm$1.10 & 20.41$\pm$1.04 & \multicolumn{1}{l}{10.41$\pm$1.03} \\
          & PubMed-SAGE(I) & 67.66$\pm$0.99 & 67.20$\pm$1.04 & 56.03$\pm$1.42 & 55.21$\pm$1.58 & \textbf{70.97$\pm$1.04} & \textbf{70.32$\pm$0.96} & 10.93$\pm$0.32 & 3.80$\pm$0.36 & 18.83$\pm$1.06 & \multicolumn{1}{l}{8.69$\pm$0.62} \\
\cmidrule{2-12}          & computers-GCN(J) & 67.80$\pm$0.62 & 66.35$\pm$0.61 & 53.01$\pm$1.39 & 51.05$\pm$1.48 & 67.66$\pm$0.85 & 61.92$\pm$1.32 & \textbf{79.89$\pm$0.67} & \textbf{71.11$\pm$0.41} & \textbf{69.77$\pm$1.85} & \multicolumn{1}{l}{\textbf{63.20$\pm$2.26}} \\
          & photo-GCN(K) & 67.47$\pm$0.65 & 66.52$\pm$0.61 & 54.55$\pm$1.38 & 53.13$\pm$1.39 & 66.44$\pm$0.67 & 59.34$\pm$1.12 & 55.88$\pm$3.68 & 47.77$\pm$3.88 & 68.76$\pm$1.27 & \multicolumn{1}{l}{61.79$\pm$1.70} \\
    \bottomrule
    \end{tabular}}%
    \end{sc}
  \label{tab:3}%
\end{table*}%

\begin{table*}[h]
  \centering
  \caption{The comparative results of loss functions A, E, I, J, and K on datasets with a class imbalance ratio of 5.}
  \begin{sc}
  \resizebox{\linewidth}{!}{
    \begin{tabular}{cl|ll|ll|ll|ll|ll}
    \toprule
    \multirow{2}[2]{*}{GNN} & \textbf{Dataset} & \multicolumn{2}{c|}{Cora} & \multicolumn{2}{c|}{CiteSeer} & \multicolumn{2}{c|}{PubMed} & \multicolumn{2}{c|}{Amazon-computers} & \multicolumn{2}{c}{Amazon-photo} \\
          & \textbf{Imbalance Ratio: 5} & \multicolumn{1}{c}{bAcc} & \multicolumn{1}{c|}{F1} & \multicolumn{1}{c}{bAcc} & \multicolumn{1}{c|}{F1} & \multicolumn{1}{c}{bAcc} & \multicolumn{1}{c|}{F1} & \multicolumn{1}{c}{bAcc} & \multicolumn{1}{c|}{F1} & \multicolumn{1}{c}{bAcc} & \multicolumn{1}{c}{F1} \\
    \midrule
    \multirow{6}[3]{*}{GCN}  & BalancedSoftmax+TAM & 74.02$\pm$0.70 & 74.08$\pm$0.67 & 61.14$\pm$1.29 & 60.41$\pm$1.31 & \textbf{74.55$\pm$0.82} & \textbf{73.92$\pm$0.99} & \textbf{84.88$\pm$0.38} & \textbf{77.21$\pm$0.22} & \underline{87.52$\pm$0.61} & \textbf{83.21$\pm$0.88} \\
\cmidrule{2-12}          & Cora-GCN(A) & \underline{75.26$\pm$0.54} & \underline{74.79$\pm$0.47} & 62.79$\pm$0.97 & 62.66$\pm$0.93 & \underline{73.61$\pm$1.23} & \underline{73.74$\pm$1.36} & 31.40$\pm$4.33 & 18.99$\pm$3.84 & 45.40$\pm$6.47 & 35.51$\pm$7.14 \\
          & CiteSeer-GAT(E) & 74.52$\pm$0.52 & 74.27$\pm$0.40 & \textbf{63.10$\pm$0.82} & \textbf{62.93$\pm$0.83} & 73.36$\pm$1.14 & 73.37$\pm$1.26 & 36.50$\pm$4.17 & 23.23$\pm$4.18 & 35.87$\pm$4.20 & 24.98$\pm$3.98 \\
          & PubMed-SAGE(I) & \textbf{75.41$\pm$0.38} & \textbf{74.84$\pm$0.48} & \underline{63.04$\pm$0.78} & \underline{62.85$\pm$0.75} & 73.31$\pm$0.94 & 73.26$\pm$0.92 & 14.29$\pm$0.68 & 6.20$\pm$0.66 & 17.73$\pm$0.80 & 10.25$\pm$1.28 \\
\cmidrule{2-12}          & computers-GCN(J) & 74.02$\pm$0.35 & 72.88$\pm$0.36 & 59.68$\pm$0.88 & 58.92$\pm$0.93 & 72.08$\pm$0.68 & 68.72$\pm$1.05 & \underline{84.63$\pm$0.37} & \underline{73.80$\pm$0.58} & \textbf{87.61$\pm$0.75} & \underline{82.77$\pm$1.20} \\
          & photo-GCN(K) & 71.47$\pm$0.54 & 71.21$\pm$0.55 & 59.31$\pm$0.96 & 58.23$\pm$1.05 & 70.18$\pm$0.79 & 65.18$\pm$1.30 & 84.22$\pm$0.29 & 73.63$\pm$0.55 & 87.03$\pm$0.83 & 81.57$\pm$1.28 \\
    \midrule
    \multirow{6}[3]{*}{GAT}  & BalancedSoftmax+TAM & 73.37$\pm$0.67 & 72.47$\pm$0.78 & 60.26$\pm$0.84 & 59.48$\pm$0.87 & \underline{74.75$\pm$0.66} & \underline{74.09$\pm$0.83} & 79.04$\pm$0.89 & \textbf{72.56$\pm$1.11} & 77.52$\pm$1.11 & \underline{73.48$\pm$1.48} \\
\cmidrule{2-12}          & Cora-GCN(A) & \textbf{76.08$\pm$0.58} & \textbf{75.37$\pm$0.55} & \textbf{64.23$\pm$0.95} & \textbf{63.84$\pm$0.93} & 73.72$\pm$0.88 & 73.48$\pm$0.82 & 44.93$\pm$7.73 & 38.84$\pm$7.07 & 77.64$\pm$6.34 & 73.06$\pm$6.70 \\
          & CiteSeer-GAT(E) & 75.34$\pm$0.33 & 74.56$\pm$0.48 & 63.58$\pm$0.76 & 63.29$\pm$0.71 & 73.83$\pm$0.90 & 73.61$\pm$0.93 & 48.14$\pm$7.60 & 41.72$\pm$6.72 & 71.81$\pm$7.11 & 66.25$\pm$7.51 \\
          & PubMed-SAGE(I) & \underline{75.87$\pm$0.42} & \underline{75.34$\pm$0.52} & \underline{63.85$\pm$0.88} & \underline{63.49$\pm$0.88} & \textbf{74.82$\pm$0.63} & \textbf{74.63$\pm$0.83} & 23.65$\pm$1.36 & 17.06$\pm$1.28 & 24.37$\pm$2.25 & 20.34$\pm$2.88 \\
\cmidrule{2-12}          & computers-GCN(J) & 61.75$\pm$1.08 & 60.77$\pm$1.26 & 49.17$\pm$0.95 & 47.93$\pm$0.91 & 69.57$\pm$0.88 & 66.43$\pm$1.47 & \underline{79.78$\pm$0.62} & 72.00$\pm$0.88 & \underline{77.79$\pm$0.52} & 72.63$\pm$0.90 \\
          & photo-GCN(K) & 65.27$\pm$0.72 & 64.72$\pm$0.82 & 49.68$\pm$1.04 & 48.31$\pm$1.20 & 71.09$\pm$0.62 & 67.56$\pm$0.94 & \textbf{80.71$\pm$0.73} & \underline{72.12$\pm$1.18} & \textbf{81.00$\pm$0.82} & \textbf{76.61$\pm$1.10} \\
              \midrule
    \multirow{6}[3]{*}{SAGE} & BalancedSoftmax+TAM & 71.71$\pm$0.58 & 71.55$\pm$0.66 & 59.55$\pm$1.29 & 58.68$\pm$1.27 & \textbf{73.36$\pm$0.55} & \textbf{72.46$\pm$0.66} & \underline{82.46$\pm$0.32} & \textbf{74.56$\pm$0.34} & \underline{82.74$\pm$1.17} & 77.18$\pm$1.68 \\
\cmidrule{2-12}          & Cora-GCN(A) & 72.92$\pm$0.58 & 72.98$\pm$0.56 & \underline{62.16$\pm$0.88} & \textbf{61.89$\pm$0.86} & 72.06$\pm$0.97 & 71.62$\pm$1.11 & 15.30$\pm$1.07 & 5.62$\pm$0.81 & 19.86$\pm$0.80 & 10.78$\pm$0.85 \\
          & CiteSeer-GAT(E) & \textbf{74.01$\pm$0.67} & \textbf{73.70$\pm$0.73} & \textbf{62.16$\pm$0.72} & \underline{61.73$\pm$0.72} & 72.25$\pm$0.70 & 71.83$\pm$0.67 & 13.33$\pm$0.65 & 4.54$\pm$0.54 & 18.80$\pm$0.46 & 9.72$\pm$0.83 \\
          & PubMed-SAGE(I) & \underline{73.93$\pm$0.62} & \underline{73.56$\pm$0.67} & 61.77$\pm$0.74 & 61.25$\pm$0.85 & \underline{72.59$\pm$0.71} & \underline{72.08$\pm$0.74} & 11.69$\pm$0.43 & 4.45$\pm$0.55 & 18.20$\pm$0.98 & 8.02$\pm$0.88 \\
\cmidrule{2-12}          & computers-GCN(J) & 73.18$\pm$0.51 & 71.61$\pm$0.61 & 58.03$\pm$1.08 & 57.31$\pm$1.10 & 71.40$\pm$0.57 & 68.70$\pm$0.71 & \textbf{82.72$\pm$0.36} & \underline{73.15$\pm$0.68} & \textbf{84.43$\pm$0.51} & \textbf{79.28$\pm$1.18} \\
          & photo-GCN(K) & 72.41$\pm$0.57 & 71.48$\pm$0.60 & 59.74$\pm$1.04 & 58.70$\pm$1.03 & 68.32$\pm$0.59 & 62.93$\pm$0.92 & 73.82$\pm$2.41 & 63.91$\pm$2.22 & 82.68$\pm$0.75 & \underline{78.31$\pm$1.23} \\
    \bottomrule
    \end{tabular}}%
    \end{sc}
  \label{tab:4}%
\end{table*}

\textbf{Q1:} Can the loss functions discovered by AutoLINC better adapt to class-imbalanced node classification tasks than SOTA alternatives? (see Section \ref{cl})

\textbf{Q2:} Are the loss functions found by AutoLINC effective across datasets with varying degrees of class imbalance? (see Section \ref{cl})

\textbf{Q3:} Can the loss functions derived from a single GNN and graph dataset demonstrate effective transferability to other GNN models and datasets? (see Section \ref{stransferability})

\textbf{Q4:} Is AutoLINC superior to SOTA non-loss function engineering methods and loss function search method? (see Section \ref{sotanonloss} and \ref{vszero})

\subsection{Experimental Setup}
\textbf{Datasets} We validate AutoLINC on well-known citation networks \cite{yang16revisiting}, comprising three datasets: Cora, CiteSeer, and PubMed, as well as Amazon's co-purchase networks \cite{mcAuley2015image}, which consist of two datasets: Computers and Photo. 

\textbf{Baseline} To evaluate the effectiveness of the loss functions learned by AutoLINC, we compare them with several baseline methods, including cross-entropy, re-weight \cite{japkowicz2002class}, balanced softmax \cite{ren2020balanced}, PC softmax \cite{hong2021disentangling}, ReNode \cite{chen2021topology}, and TAM \cite{song2022tam}. All experiments are conducted ten times to calculate bAcc and F1 scores.

\textbf{GNN Settings} Prominent GNNs: GCN \cite{kipf2017semisupervised}, GAT \cite{veličković2018graph}, and SAGE \cite{hamilton2017inductive}, consist of a 2-layer neural network with a hidden layer dimension of 256. More details on datasets, baselines and parameters are shown in Appendix \ref{es}.

\subsection{Comparison with SOTA Loss Functions}
\label{cl}
The outcomes of AutoLINC and SOTA loss functions are detailed in Tables \ref{tab:1} and \ref{tab:2}. The discovered loss functions A-O and their convergence trends are presented in Appendix Table \ref{tab:exp} and Appendix Fig. \ref{fig:lc}, respectively. Our observations are as follows: 1) AutoLINC consistently demonstrates significant performance enhancements, with the most substantial gains seen with GAT and SAGE, while GCN experiences a more modest improvement. 2) AutoLINC leads to remarkable performance improvements on datasets such as Cora, CiteSeer, Amazon-computers, and Amazon-photo. However, the enhancement on PubMed is moderate, possibly due to its high degree of topology imbalance.

\subsection{Transferability and Convergence}
\label{stransferability}
\textbf{Transferability to Different Imbalance Ratios}.  
We additionally examined the transferability of the discovered loss functions from datasets with an initial imbalance ratio of $\rho=10$ to datasets with $\rho=5$, as outlined in Table \ref{tab:2}. Impressively, these loss functions retained their high level of performance, underlining their robustness in facilitating node classification across varying class imbalance ratios.

\begin{table*}[htbp]
\tiny
  \centering
  \caption{The performance of AutoLINC trained on three GNNs. Here, $\rho=10$. AutoLINC-Cora: $(\tanh(-((1/N \times (-y)+\hat{y})))^2$; AutoLINC-CiteSeer: $(\hat{y}+((1/N \times (-y))+\hat{y}))^2$; AutoLINC-PubMed: $N\times \hat{y}^4+\left(-\left(\hat{y} \times y\right)\right)$.}
  \vskip -0.1in
  \begin{sc}
    \begin{tabular}{c|l|cc|cc|cc}
    \toprule
    \multirow{2}[2]{*}{GNN} & Dataset & \multicolumn{2}{c|}{Cora} & \multicolumn{2}{c|}{CiteSeer} & \multicolumn{2}{c}{PubMed} \\
          & Imbalance Ratio: 10 & bAcc  & F1    & bAcc  & F1    & bAcc  & F1 \\
    \midrule
    \multirow{4}[2]{*}{GCN} & BalancedSoftmax+TAM & 69.17$\pm$0.77 & 69.00$\pm$0.73 & 55.65$\pm$1.19 & 54.06$\pm$1.39 & \underline{72.15$\pm$1.08} & \underline{71.79$\pm$1.35} \\
          & AutoLINC-Cora & \textbf{69.63$\pm$0.77} & 68.95$\pm$0.79 & \underline{57.52$\pm$1.76} & \textbf{57.14$\pm$1.82} & 71.82$\pm$1.20 & 71.27$\pm$1.41 \\
          & AutoLINC-CiteSeer & \underline{69.55$\pm$0.60} & \underline{69.06$\pm$0.61} & \textbf{57.78$\pm$1.63} & \underline{57.10$\pm$1.71} & 72.09$\pm$1.54 & 71.29$\pm$1.60 \\
          & AutoLINC-PubMed & 69.07$\pm$0.61 & \textbf{69.23$\pm$0.51} & 55.48$\pm$1.81 & 54.91$\pm$1.81 & \textbf{73.37$\pm$1.03} & \textbf{72.11$\pm$1.43} \\
    \midrule
    \multirow{4}[2]{*}{GAT} & BalancedSoftmax+TAM & 66.30$\pm$1.01 & 65.28$\pm$1.03 & 54.14$\pm$1.31 & 51.84$\pm$1.63 & \underline{72.24$\pm$0.85} & \textbf{71.96$\pm$0.76} \\
          & AutoLINC-Cora & \underline{70.55$\pm$0.96} & \underline{69.97$\pm$1.02} & 58.31$\pm$1.61 & \underline{57.40$\pm$1.64} & 72.23$\pm$1.09 & 70.98$\pm$1.26 \\
          & AutoLINC-CiteSeer & 70.13$\pm$0.82 & 69.89$\pm$0.98 & \textbf{59.15$\pm$1.43} & \textbf{58.11$\pm$1.65} & 72.22$\pm$1.19 & 71.72$\pm$1.41 \\
          & AutoLINC-PubMed & \textbf{71.41$\pm$0.62} & \textbf{71.66$\pm$0.67} & \underline{58.51$\pm$1.45} & 57.21$\pm$1.75 & \textbf{73.00$\pm$0.96} & \underline{71.80$\pm$1.12} \\
    \midrule
    \multirow{4}[2]{*}{SAGE} & BalancedSoftmax+TAM & 66.54$\pm$0.49 & 66.24$\pm$0.65 & 52.46$\pm$1.23 & 49.27$\pm$1.72 & 69.64$\pm$0.93 & \underline{70.26$\pm$0.87} \\
          & AutoLINC-Cora & \textbf{68.73$\pm$0.92} & \textbf{68.72$\pm$0.89} & \underline{56.34$\pm$1.40} & \underline{55.10$\pm$1.65} & 69.68$\pm$0.90 & 68.76$\pm$1.15 \\
          & AutoLINC-CiteSeer & \underline{68.47$\pm$0.99} & \underline{68.43$\pm$1.02} & \textbf{56.54$\pm$1.14} & \textbf{55.45$\pm$1.36} & \underline{69.98$\pm$0.96} & 69.31$\pm$1.17 \\
          & AutoLINC-PubMed & 67.27$\pm$0.71 & 67.29$\pm$0.70 & 54.76$\pm$1.60 & 53.66$\pm$1.88 & \textbf{71.50$\pm$0.67} & \textbf{70.42$\pm$1.04} \\
    \bottomrule
    \end{tabular}%
    \end{sc}
  \label{tab:5}%
\end{table*}%

\begin{table*}[h]
  \centering
  \begin{sc}
  \caption{Comparison with SOTA non-loss function
engineering methods.}
\vskip -0.1in
    \resizebox{\linewidth}{!}{
    \begin{tabular}{c|l|cccc|cccc|cccc}
    \toprule
    \multirow{3}{*}{\centering GNN} & Dataset & \multicolumn{4}{c|}{Cora} & \multicolumn{4}{c|}{CiteSeer} & \multicolumn{4}{c}{PubMed} \\
          & Imbalance Ratio & \multicolumn{2}{c}{5}& \multicolumn{2}{c|}{10} & \multicolumn{2}{c}{5}& \multicolumn{2}{c|}{10} & \multicolumn{2}{c}{5}& \multicolumn{2}{c}{10}\\
          & ~ & bAcc  & F1 & bAcc  & F1  & bAcc  & F1 & bAcc  & F1   & bAcc  & F1 & bAcc  & F1\\
    \midrule
    \multirow{5}{*}{\centering GCN} & GraphSMOTE & 69.53$\pm$0.85 & 69.17$\pm$1.02 & 65.01$\pm$0.95 & 64.40$\pm$1.08 & 50.50$\pm$0.71 & 47.16$\pm$0.98 & 41.75$\pm$1.45 & 36.71$\pm$2.16 & 67.83$\pm$0.73 & 65.53$\pm$1.21 & 64.90$\pm$1.04 & 61.65$\pm$1.95 \\
          & GraphENS & 75.68$\pm$0.39 & 75.51$\pm$0.57 & 70.03$\pm$0.83 & 69.98$\pm$0.89 & \underline{62.81$\pm$0.82} & \underline{62.25$\pm$0.86} & 54.32$\pm$1.53 & 51.53$\pm$2.20 & 75.09$\pm$0.66 & 73.92$\pm$0.91 & \underline{72.59$\pm$1.27} & 70.44$\pm$1.63  \\
          & GraphSHA & \underline{76.69$\pm$0.32} & \textbf{76.45$\pm$0.36} & \underline{73.37$\pm$0.60} & \textbf{73.09$\pm$0.65} & 62.50$\pm$1.69 & 61.81$\pm$1.82 & 56.42$\pm$1.88 & 52.97$\pm$2.26 & \textbf{75.40$\pm$0.64} & 73.81$\pm$0.69 & 72.35$\pm$0.98 & 69.66$\pm$1.37 \\
          & AutoLINC & 75.26$\pm$0.54 & 74.79$\pm$0.47 & 70.21$\pm$0.67 & 69.67$\pm$0.79 & 60.64$\pm$1.05 & 60.39$\pm$1.01 & \underline{56.56$\pm$1.97} & \underline{56.09$\pm$2.06} & 74.42$\pm$0.86 & \textbf{74.61$\pm$0.88} & 72.43$\pm$1.03 & \underline{72.42$\pm$1.07} \\
          & AutoLINC(GraphSHA) & \textbf{76.88$\pm$0.43} & \underline{76.24$\pm$0.40} & \textbf{73.75$\pm$0.77} & \underline{72.93$\pm$0.86} & \textbf{63.10$\pm$1.31} & \textbf{62.59$\pm$1.39} & \textbf{58.57$\pm$1.88} & \textbf{56.68$\pm$2.25} & \underline{75.27$\pm$0.58} & \underline{74.53$\pm$0.65} & \textbf{73.32$\pm$1.10} & \textbf{72.74$\pm$0.91} \\
    \midrule
    \multirow{5}{*}{\centering GAT} & GraphSMOTE & 71.84$\pm$0.55 & 71.44$\pm$0.55 & 68.93$\pm$0.86 & 68.16$\pm$0.97 & 60.66$\pm$1.04 & 59.73$\pm$1.43 & 53.94$\pm$2.59 & 52.52$\pm$3.14 & 69.96$\pm$1.27 & 67.11$\pm$1.72 & 67.93$\pm$0.90 & 63.28$\pm$1.58\\
          & GraphENS & 75.18$\pm$0.68 & 74.39$\pm$0.68 & 70.02$\pm$0.94 & 68.99$\pm$1.00 & 58.94$\pm$1.63 & 57.79$\pm$1.82 & 52.13$\pm$1.56 & 48.54$\pm$2.07 & 74.10$\pm$0.80 & 72.85$\pm$1.04 & 71.22$\pm$1.04 & 68.22$\pm$1.39 \\
          & GraphSHA & 73.07$\pm$0.21 & 73.40$\pm$0.33 & 68.16$\pm$1.00 & 68.31$\pm$1.14 & 60.47$\pm$1.87 & 58.95$\pm$2.33 & 53.64$\pm$2.61 & 48.84$\pm$3.22 & 74.00$\pm$0.92 & 72.56$\pm$1.11 & 71.56$\pm$1.16 & 68.33$\pm$1.65 \\
          & AutoLINC & \underline{75.57$\pm$0.42} & \underline{75.24$\pm$0.40} & \underline{70.63$\pm$1.01} & \underline{70.16$\pm$1.04} & \textbf{63.58$\pm$0.76} & \textbf{63.29$\pm$0.71} & \textbf{58.48$\pm$1.57} & \textbf{57.38$\pm$1.89} & \underline{74.51$\pm$0.82} & \textbf{74.50$\pm$0.79} & \underline{72.57$\pm$0.88} & \underline{72.27$\pm$1.02}  \\
          & AutoLINC(GraphSHA) & \textbf{75.85$\pm$0.55} & \textbf{75.76$\pm$0.58} & \textbf{71.97$\pm$0.74} & \textbf{71.57$\pm$0.76} &  \underline{61.97$\pm$1.21} & \underline{61.44$\pm$1.26} & \underline{57.26$\pm$1.74} & \underline{55.07$\pm$2.11} & \textbf{74.86$\pm$0.54} & \underline{74.22$\pm$0.71} & \textbf{73.71$\pm$0.87} & \textbf{72.60$\pm$0.87} \\
    \midrule
    \multirow{5}{*}{\centering SAGE} & GraphSMOTE & 63.64$\pm$1.03 & 63.12$\pm$1.19 & 58.53$\pm$1.01 & 56.59$\pm$1.35 & 51.93$\pm$1.22 & 50.39$\pm$1.33 & 44.52$\pm$2.00 & 40.15$\pm$2.66 & 72.02$\pm$0.32 & 69.50$\pm$0.65 & 68.42$\pm$0.88 & 62.97$\pm$1.23\\
          & GraphENS & 72.35$\pm$0.53 & 72.39$\pm$0.70 & 66.58$\pm$0.74 & 66.35$\pm$0.83 & 61.11$\pm$0.77 & 60.63$\pm$0.77 & 53.91$\pm$1.24 & 52.23$\pm$1.51 & 72.90$\pm$0.62 & 72.59$\pm$0.69 & 71.34$\pm$0.79 & 69.56$\pm$0.96 \\
          & GraphSHA & \underline{75.74$\pm$0.44} & \underline{75.24$\pm$0.54} & \underline{72.47$\pm$0.52} & \underline{71.91$\pm$0.60} & \textbf{62.09$\pm$1.53} & \textbf{61.81$\pm$1.57} & \underline{57.01$\pm$1.85} & \underline{53.95$\pm$2.31} & \underline{74.48$\pm$0.53} & \underline{73.91$\pm$0.62} & \underline{72.84$\pm$0.81} & \underline{70.79$\pm$1.19} \\
          & AutoLINC & 74.15$\pm$0.69 & 73.80$\pm$0.70 & 68.51$\pm$1.02 & 68.49$\pm$1.02 & \underline{61.41$\pm$0.85} & \underline{61.13$\pm$0.78} & 56.78$\pm$1.27 & \textbf{56.16$\pm$1.37} & 72.59$\pm$0.71 & 72.08$\pm$0.74 & 70.97$\pm$1.04 & 70.32$\pm$0.96 \\
          & AutoLINC(GraphSHA) & \textbf{76.76$\pm$0.53} & \textbf{76.07$\pm$0.47} & \textbf{73.50$\pm$0.69} & \textbf{72.72$\pm$0.83} & 61.09$\pm$1.56 & 59.88$\pm$1.94 & \textbf{57.03$\pm$1.80} & 53.86$\pm$2.27 & \textbf{75.44$\pm$0.38} & \textbf{75.11$\pm$0.31} & \textbf{73.29$\pm$1.12} & \textbf{72.37$\pm$1.29} \\
    \bottomrule
    \end{tabular}}%
  \label{tab:6}%
  \end{sc}
\end{table*}%

\textbf{Transferability across different datasets and GNNs}. 
We evaluated the transferability of five distinct loss functions, indicated as A, E, I, J, and K (as listed in Appendix Table \ref{tab:exp}), in various combinations of data sets and graph neural networks, as presented in Table \ref{tab:3}. Our observations are: 1) Loss functions A, E, and I demonstrate robust transferability within datasets of the same type (citation networks) and even surpass other loss functions in certain scenarios. 2) Likewise, J and K exhibit excellent performance when applied to datasets of the same type (Amazon co-purchase networks). 3) However, when these loss functions are transferred to datasets of different types, their transfer performance diminishes. Notably, A, E, and I, trained on citation network datasets, exhibit limited classification performance on Amazon datasets.

Consequently, we can conclude that homophily in graph-structured data significantly enhances AutoLINC's transferability. Additionally, as detailed in Table \ref{tab:4}, as the class imbalance ratio is reduced to 5, models trained using the loss functions from Table \ref{tab:exp} display enhanced classification capabilities due to the decreased class imbalance. Similarly, they continue to exhibit superior transfer performance within datasets of the same type, but their transfer performance diminishes. In some instances, they exhibit no transferability to datasets of different types.

\textbf{Convergence}. In Appendix Fig. \ref{fig:lc}, we present the average bAcc curves for the discovered loss functions detailed in Appendix Table \ref{tab:exp}. Although the convergence is relatively slow, the final model outperforms other loss functions in most cases. Additionally, Table \ref{tab:1} illustrates the highly competitive performance of the best-trained model.

\textbf{Discussion}. 
We further investigate the potential performance enhancements achieved by AutoLINC across various GNNs. Unlike Eq. \ref{eq:1}, we employ the lowest score among the three scores obtained from GCN, GAT, and SAGE models for a particular dataset to evaluate the loss function. In Table \ref{tab:5}, we observe: 1) AutoLINC demonstrates superior performance compared to loss functions discovered within the scope of a single GNN model; 2) This strategy consistently delivers enhanced classification performance when compared to state-of-the-art loss functions; 3) However, the loss functions generated by this approach exhibit reduced transferability across different GNNs when compared to loss functions A, E, and I.

Our method requires additional training and evaluation by training the model on the training set using the learned loss function and evaluating the model on the validation set to obtain scores for the loss function. Methods based on loss functions train the model on the training set, which is similar to our approach in terms of the use of the training set. Additionally, the training set under the adopted class imbalance setting is also small. Our method does not heavily rely on the training set. The impact on the validation set is that our proxy task requires evaluating the score of the loss function on the validation set to guide AutoLINC in searching for the loss function. If the validation set is too small, it may lead to the search framework discovering loss functions overfitting the validation set. To investigate, we conducted experiments to evaluate on smaller validation sets. We used three different combinations of datasets and networks, namely Cora-GCN, CiteSeer-GAT, and PubMed-GraphSAGE. The training and testing sets remained unchanged. The size of the validation set was set to maintain 5, 10, and 20 nodes per class. The compared methods were Balanced Softmax and TAM, and we did not perform any tuning for any method. Other settings were the same as in Table \ref{tab:1}. The experimental results are presented in Table \ref{tab:valsize}.

If reducing the size of the validation set leads to a decrease in the performance of node classification, it may be due to overfitting of the loss function found by our method on the validation set, or it may be because the selected best model is not optimal, or due to the deterioration of other model parameters. This is difficult to analyze. Therefore, the conclusions drawn from experiments with different validation set sizes may not be convincing. However, by comparing with Balanced Softmax and TAM under the same validation set size settings, our method still demonstrates good performance on smaller validation sets. This suggests that our method does not heavily rely on the validation set.

\begin{table*}[!ht]
    \centering
    \begin{sc}
    \caption{Experimental results under different sizes of validation set. Here, we set it to 5, 10, and 20 nodes per class.}
    \vskip -0.1in
    \resizebox{\linewidth}{!}{
    \begin{tabular}{l|l|cc|cc|cc}
    \toprule
        Datasets  & Methods & \multicolumn{2}{c|}{5} & \multicolumn{2}{c|}{10} & \multicolumn{2}{c}{20} \\ 
        \& GNNs & ~ & bACC & F1 & bACC & F1 & bACC & F1 \\ \midrule
        \multirow{3}{*}{\centering Cora-GCN} & BalancedSoftmax & 67.48$\pm$1.04 & 66.20$\pm$1.28 & 68.96$\pm$0.55 & \textbf{68.67$\pm$0.52} & 68.96$\pm$0.55 & \textbf{68.73$\pm$0.51} \\ 
        ~ & BalancedSoftmax+TAM & 66.98$\pm$1.41 & 65.88$\pm$1.55 & 67.48$\pm$1.40 & 66.37$\pm$1.59 & 68.76$\pm$0.91 & 68.73$\pm$1.02 \\ 
        ~ & AutoLINC & \textbf{68.65$\pm$1.26} & \textbf{67.33$\pm$1.13} & \textbf{69.08$\pm$0.87} & 67.80$\pm$0.71 & \textbf{69.51$\pm$0.97} & 67.30$\pm$1.07 \\ 
        \midrule
        \multirow{3}{*}{\centering CiteSeer-GAT} & BalancedSoftmax & 51.18$\pm$1.82 & 47.73$\pm$2.57 & 51.08$\pm$1.32 & 48.84$\pm$1.79 & 53.70$\pm$1.64 & 50.31$\pm$2.03 \\ 
        ~ & BalancedSoftmax+TAM & 51.00$\pm$1.49 & 47.98$\pm$2.31 & 51.70$\pm$1.69 & 48.48$\pm$2.37 & 53.51$\pm$1.64 & 51.99$\pm$1.83 \\ 
        ~ & AutoLINC & \textbf{54.46$\pm$2.27} & \textbf{53.18$\pm$2.47} & \textbf{54.69$\pm$2.43} & \textbf{53.32$\pm$2.76} & \textbf{55.39$\pm$2.24} & \textbf{53.61$\pm$2.71} \\ 
        \midrule
        \multirow{3}{*}{\centering PubMed-SAGE} & BalancedSoftmax & 68.98$\pm$0.76 & 65.57$\pm$1.21 & 69.09$\pm$0.74 & 66.78$\pm$1.16 & 69.77$\pm$0.73 & 68.63$\pm$0.87 \\ 
        ~ & BalancedSoftmax+TAM & \textbf{69.26$\pm$0.91} & \textbf{66.89$\pm$1.38} & \textbf{69.99$\pm$0.71} & \textbf{68.53$\pm$1.03} & 69.23$\pm$0.65 & \textbf{69.10$\pm$0.73} \\ 
        ~ & AutoLINC & 67.54 $\pm$ 0.83 & 65.78 $\pm$ 1.26 & 69.58$\pm$0.79 & 67.81$\pm$1.34 & \textbf{70.14$\pm$0.47} & 68.20$\pm$1.02 \\ 
        \bottomrule
    \end{tabular}}
    \label{tab:valsize}
    \end{sc}
\end{table*}

\subsection{Comparison with SOTA Non-loss Function Engineering Methods}
\label{sotanonloss}
To enhance the generality of the baseline, we have included comparative experiments with SOTA non-loss function engineering, GraphSMOTE \cite{GraphSMOTE}, GraphENS \cite{park2022graphens}, and GraphSHA \cite{GraphSHA} see Appendix \ref{es} for more information). 
The experimental setting is the same as in Section \ref{cl}. The results of AutoLINC derived from Table \ref{tab:1} and \ref{tab:2} are competitive to GraphSHA. We also combine AutoLINC with GraphSHA to discover loss functions so as to explore the scalability of AutoLINC. In Table \ref{tab:6}, in comparison to GraphSMOTE, GraphENS and GraphSHA, AutoLINC combined with GraphSHA demonstrates advanced performancere across different datasets, GNN backbones, and class imbalance ratios. This illustrates the scalability of AutoLINC and the feasibility of overcoming the challenges of class imbalance in node classification from a loss function standpoint.

\subsection{AutoLINC v.s. Autoloss-Zero}
\label{vszero}
We demonstrate the advantages of AutoLINC over Autoloss-Zero \cite{li2021autoloss}. 
AutolossZero-A adopts the original search space of AutolossZero, modified to accommodate graph data by excluding some inapplicable primitive operators. Correspondingly, its loss function rejection strategies and proxy task align with AutoLINC to ensure fairness. AutolossZero-B is a variant that introduces the parameter $N$ into the loss functions within its search space, while the rest remains unchanged. Under the same time constraints (1 hour CPU time), experimental results are presented in Table \ref{non-loss}. With the exception of differences in the search method and a search depth set to 4, all other aspects remain consistent with our AutoLINC method. AutoLINC achieves better performance than Autoloss-Zero. This is attributed to AutoLINC's inclination to discover structurally simple trees in proxy tasks, whereas AutolossZero tends to find complex loss functions. In the task of node classification, structurally simple trees are more likely to exhibit good performance in overall task evaluation.

\begin{table*}[ht]
    \centering
    \caption{The comparative results of loss functions searched by different frameworks.}
    \vskip -0.1in
    \resizebox{\linewidth}{!}{
    \begin{sc}
        \begin{tabular}{c|l|cc|cc|cc}
        \toprule
        \multirow{2}[2]{*}{GNN} & Dataset & \multicolumn{2}{c|}{Cora} & \multicolumn{2}{c|}{CiteSeer} & \multicolumn{2}{c}{PubMed} \\
          &  Imbalance Ratio:10 & bAcc & F1 & bAcc & F1 & bAcc & F1 \\
        \midrule
        \multirow{3}[2]{*}{GCN} & Autoloss-Zero-A & 55.57$\pm$1.02 & 54.71$\pm$1.08 & 39.63$\pm$1.07 & 33.02$\pm$1.74 & 65.79$\pm$1.61 & 57.17$\pm$3.43 \\
        & Autoloss-Zero-B & 62.64$\pm$0.89 & 61.58$\pm$1.21 & 55.73$\pm$1.67 & 54.29$\pm$1.94 & 72.41$\pm$1.08 & 71.71$\pm$1.19 \\
        & AutoLINC(1 hour) & \textbf{70.60$\pm$0.62} & \textbf{69.13$\pm$0.53} & \textbf{56.63$\pm$1.97} & \textbf{55.98$\pm$2.05} & \textbf{72.90$\pm$0.71} & \textbf{72.56$\pm$0.86} \\
        \midrule
        \multirow{3}[2]{*}{GAT} & Autoloss-Zero-A & 63.07$\pm$0.84 & 60.13$\pm$0.89 & 46.44$\pm$0.99 & 41.05$\pm$1.47 & \textbf{69.43$\pm$1.19} & \textbf{65.85$\pm$2.48} \\
        & Autoloss-Zero-B & 68.03$\pm$0.49 & 67.67$\pm$0.53 & 55.61$\pm$1.24 & 55.50$\pm$1.12 & 65.82$\pm$1.77 & 64.65$\pm$1.57 \\
        & AutoLINC(1 hour) & \textbf{71.34$\pm$0.90} & \textbf{71.16$\pm$0.91} & \textbf{57.78$\pm$1.62} & \textbf{57.19$\pm$1.76} & 63.69$\pm$1.69 & 62.65$\pm$2.27 \\
        \midrule
        \multirow{3}[2]{*}{SAGE} & Autoloss-Zero-A & 55.34$\pm$0.32 & 51.30$\pm$0.56 & 43.66$\pm$0.77 & 36.78$\pm$1.21 & 65.80$\pm$1.82 & 64.71$\pm$1.94 \\
        & Autoloss-Zero-B & 68.68$\pm$0.86 & \textbf{68.37$\pm$0.87} & 42.63$\pm$1.63 & 37.83$\pm$1.91 & \textbf{72.46$\pm$0.60} & \textbf{70.49$\pm$0.88} \\
        & AutoLINC(1 hour) & \textbf{68.92$\pm$0.82} & 67.99$\pm$0.90 & \textbf{56.59$\pm$1.41} & \textbf{55.56$\pm$1.53} & 71.08$\pm$0.88 & 70.02$\pm$0.96 \\ \bottomrule
    \end{tabular}
    \end{sc}
    }
    \label{non-loss}
    \vskip -0.1in
\end{table*}

\subsection{Ablation Study}
\begin{figure}[htbp]
    \centering
    \includegraphics[width=0.8\linewidth]{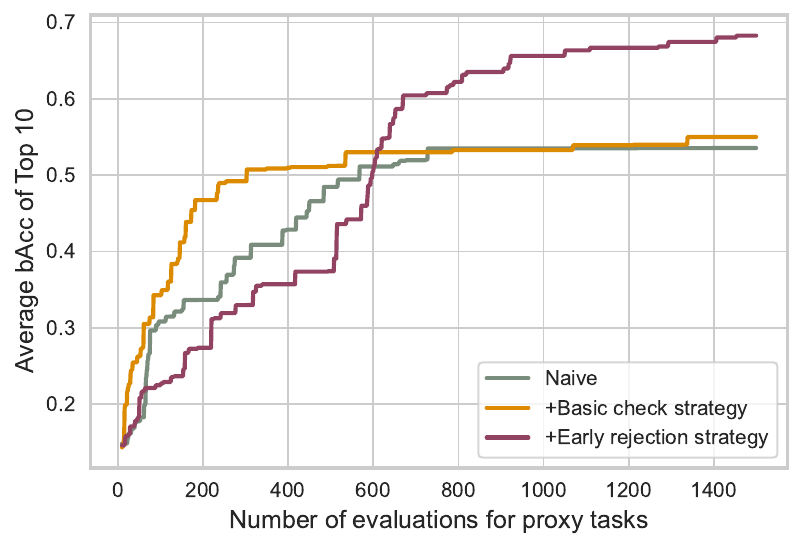}
    \caption{This figure illustrates the average scores of the Top 10 loss metrics during the search process. Naive represents MCTS without the proposed strategies.}
    \label{fig:as}
\end{figure}

The ablation studies were conducted on Cora using the GCN model, with a total of 10 trials. In Figure \ref{fig:as}, we showcase the contrasting impacts of the Basic Check Strategy and the Early Rejection Strategy, both essential components of our approach. 
The Basic Check strategy consistently outperforms the Naive approach within the same number of evaluations, thus expediting the search process.
The adaptability of the rejection threshold, which aligns with the Top 10, ensures that as the search progresses, only loss functions that converge rapidly and demonstrate superior performance are retained.

Further insight into the efficiency of AutoLINC is provided in Table \ref{tab:as}, which details the number of valid loss functions explored within a 12-hour timeframe. The integration of the Basic Check Strategy accelerates the search process by nearly 6-fold, while the addition of the Early Rejection Strategy enhances search efficiency by almost 8-fold. These findings underscore the substantial improvements in the search efficiency of AutoLINC due to the incorporation of these two loss-checking strategies.

\begin{table}[htbp]
    \centering
    \caption{AutoLINC explores the number of valid loss functions within 12 hours using different strategy combinations.}
    \begin{sc}
    \resizebox{\linewidth}{!}{
    \begin{tabular}{cccc}
    \toprule
Basic Check & Early Rejection	& No. 
Loss	& Speed-Up \\
Strategy & Strategy	&  
Functions	&  \\
\midrule
\XSolidBrush & \XSolidBrush &	$8.3\times10^4$ & $1\times$ \\
\Checkmark & \XSolidBrush &		$4.8\times10^5$ &	$5.7\times$\\
\Checkmark & \Checkmark &	$6.5\times10^5$	& $7.8\times$\\
\bottomrule
    \end{tabular}
    }
    \end{sc}
    \label{tab:as}
\end{table}

\begin{figure}[htb]
\vskip -0.2in
\centering
\includegraphics[width=0.85\linewidth]{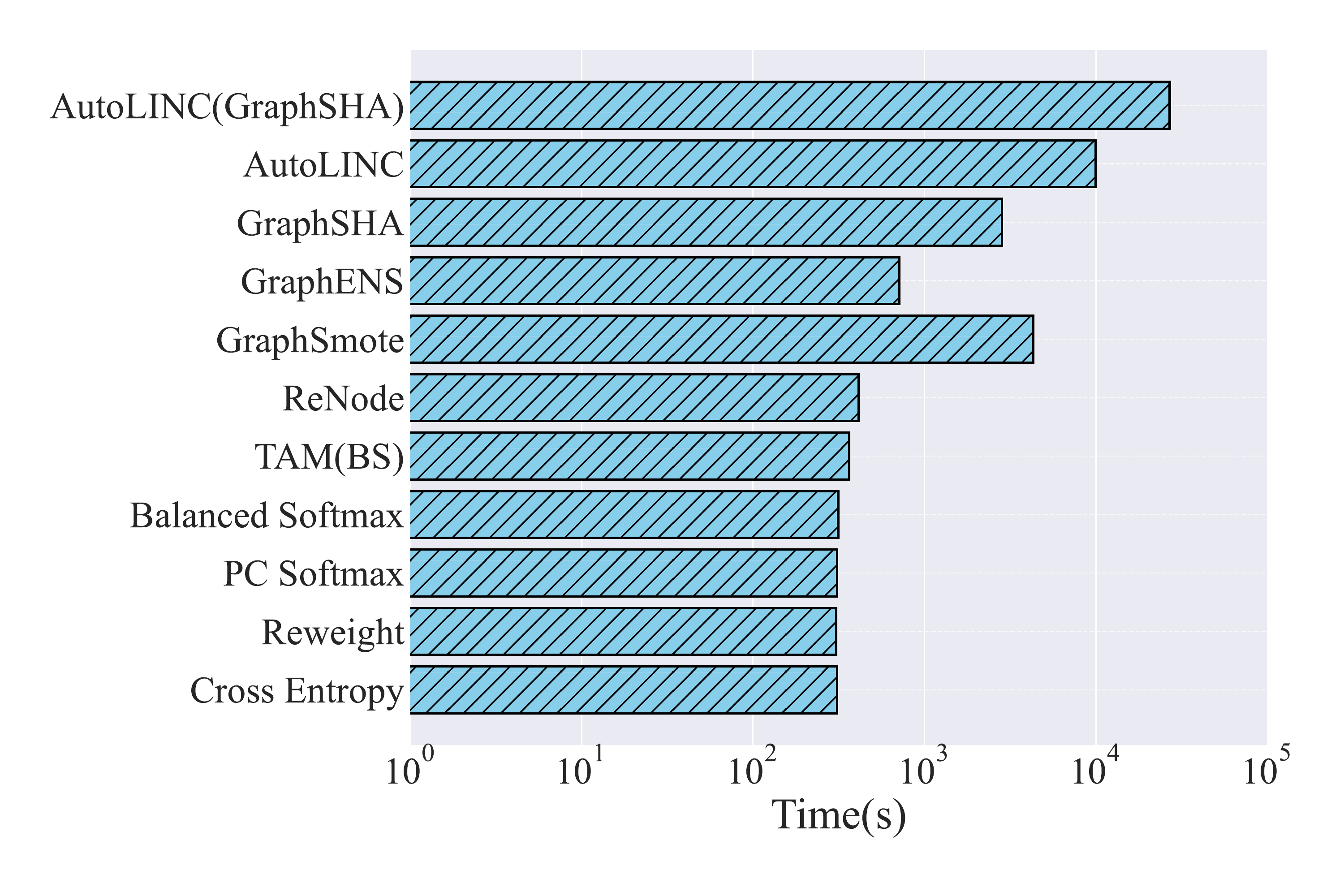}\\
\vskip -0.2in
\caption{The runtime on GCN across PubMed datasets using 10 random seeds.}
\vskip -0.2in
\label{fig:time}
\end{figure}

\textbf{Time}
In Figure \ref{fig:time}, AutoLINC exhibits the highest time consumption among the compared methods. However, this time consumption is justified by the ratio of performance gain achieved. Additionally, the discovered loss functions demonstrate a degree of transferability, suggesting the potential for addressing the problem with reduced time costs.

\section{Conclusions}
This paper introduces an automatic loss function search framework with high performance and generalization capabilities for addressing class-imbalanced node classification problems. Compared to SOTA loss functions, the functions discovered within this framework demonstrate significant improvements in classification performance, affirming the effectiveness of the proposed framework. We also find that the loss functions discovered based on a single GNN and dataset exhibit transferability to homogeneous datasets. Crucially, they compete favorably with SOTA loss functions. Furthermore, we observe that loss functions discovered under high class-imbalance ratios generalize well to scenarios with lower class-imbalance ratios, highlighting the adaptability of our proposed approach. Finally, we validate that the employed Basic Check and Early Rejection Strategies can accelerate the operation of the search algorithm.
Further research is warranted to design an Autoloss framework with transferability on heterogeneous graph.

\section*{Acknowledgements}
This work was supported in part by the National Natural Science Foundation of China under Grant 62206205, in part by the Young Talent Fund of Association for Science and Technology in Shaanxi, China under Grant 20230129, in part by the Guangdong High-level Innovation Research Institution Project under Grant 2021B0909050008, and in part by the Guangzhou Key Research and Development Program under Grant 202206030003.

\section*{Impact Statement}
This paper presents work whose goal is to advance the field of Machine Learning. There are many potential societal consequences of our work, none which we feel must be specifically highlighted here.




\newpage
\appendix
\onecolumn

\begin{algorithm}[htbp]
\small
\caption{AutoLINC}
\label{alg:1}
\begin{algorithmic}
   \STATE {\bfseries Input:} Grammar $\mathcal{G}=(U, \Sigma, R, O)$, node feature matrix $X$, node label $Y$, GNN model $f_\theta$, measure\ $\sigma$, number of trails $P$, number of episode $EP$, number of simulation $B$, $t_{max}$\;
   \STATE {\bfseries Output:} Optimal loss function $\mathcal{L}^\star$\; 
   \STATE Initialize $M=\emptyset$\;
    \FOR{$p\leftarrow 1, P$}
        \FOR{$e \leftarrow 1, EP$}
        \STATE \textbf{Selection:} Initialize $s_0=\emptyset, t=0, NT=[O]$\;
            \WHILE {$s_t$ expandable and $t < t_{max}$}
           \STATE $a_{t+1} \leftarrow \arg \max_{a \in \mathcal{A}} UCT(s_t, a)$\; 
           \STATE Obtain $s_{t+1},NT$ after $a_{t+1}$; $t\leftarrow t+1$\;
            \ENDWHILE
         \STATE \textbf{Expansion:} Randomly take an unvisited path with $a$\;
        Obtain $s_{t+1},NT$ after $a$; $t\leftarrow t+1$\;
        \IF{$NT = \emptyset$}
           \STATE Validate the Legitimacy of $\mathcal{L}$ and check $\mathcal{L}$ via the Loss Check Strategy\;
           \STATE Train the proxy task with $\mathcal{L}$\;
           \STATE $r\leftarrow \sigma(f_\theta(X_{val},Y_{val}))$\;
           \STATE Record $\mathcal{L}, r$ in $ M$, \textbf{Backpropagate} and finish the episode\;
        \ENDIF
        \STATE \textbf{Simulation:} Fix the state $s_t$ and $NT$\;
        Initialize $r=0$\;
        \FOR{$b\leftarrow 1, B$}
        \WHILE {$s_t$ non-terminal and $t < t_{max}$}
       \STATE $a\leftarrow Random(\mathcal{A})$\;
       \STATE Obtain $s_{t+1},NT$ after $a_{t+1}$; $t\leftarrow t+1$\;
        \ENDWHILE
        \IF{$NT = \emptyset$}
           \STATE Validate the Legitimacy of $\mathcal{L}$ and check \STATE $\mathcal{L}$ via the Loss Check Strategy\;
           \STATE Train the proxy task with $\mathcal{L}$\;
           \STATE $r\leftarrow \sigma(f_\theta(X_{val}, Y_{val}))$\;
           \STATE Record $\mathcal{L}, r$ in $M$\;
        \ENDIF
        \ENDFOR
        \STATE \textbf{Backpropagate:} Back-update parent nodes based on simulation results all the way back to the root node\;
        \ENDFOR
    \ENDFOR
    \STATE Get the top-10 loss functions $M_{top}$ from $M$ based on reward\;
    \STATE Initialize $best=0$, $\mathcal{L}^{*}=\emptyset$\;
    \FOR{$l\leftarrow 1, 10$}
    \STATE $\mathcal{L}\leftarrow M_{top}(l)$\;
    \STATE Train the complete task with $\mathcal{L}$\;
    \STATE $r\leftarrow \sigma(f_\theta(X_{val}, Y_{val}))$\;
    \IF{$r>best$}
    \STATE $best\leftarrow r$;
    \STATE $\mathcal{L}^\star\leftarrow \mathcal{L}$\;
    \ENDIF
    \ENDFOR
    \end{algorithmic}
\end{algorithm}

\section{Experimental Setup}
\label{es}
\textbf{Datasets} We validate AutoLINC on well-known citation networks \cite{yang16revisiting}, comprising three datasets: Cora, CiteSeer, and PubMed, as well as Amazon's co-purchase networks \cite{mcAuley2015image}, which consist of two datasets: Computers and Photo. In the case of citation networks, we employ training, validation, and testing set splits as described in \cite{yang16revisiting}. For Amazon networks, the data set is divided into five distinct partitions using five different random seeds, following the methodology in \cite{chen2021topology}. A label ratio of 0.01 is maintained. To generate class-imbalanced datasets, we adopt the step imbalance method as detailed in \cite{park2022graphens,GraphSMOTE}. The minority class contains an equal number of instances, denoted as $n$, while the majority class consists of $n\times \rho$ instances, where $\rho$ represents the class imbalance ratio. For this experiment, we set $\rho$ to 5 or 10.

\textbf{Baseline} To evaluate the effectiveness of the loss functions learned by AutoLINC, we compare them with several baseline methods, including cross-entropy, re-weight \cite{japkowicz2002class}, balanced softmax \cite{ren2020balanced}, PC softmax \cite{hong2021disentangling}, ReNode \cite{chen2021topology}, and TAM \cite{song2022tam}. Cross-entropy and re-weight serve as fundamental baseline approaches, while balanced softmax and PC softmax are designed to address long-tail issues in computer vision. On the other hand, ReNode and TAM are specifically tailored to tackle imbalanced node classification problems. For ReNode, we enhance it by combining it with the focal loss, with a focal hyperparameter set to 2.0, and topology imbalance bounds set to 0.5 and 1.5. In the case of TAM, we carefully select parameters based on the average of accuracy and F1 score from the recommended settings in its original paper. These parameters include the Anomalous Connectivity Margin term coefficients ($\{0.25, 0.5, 1.5, 2.5\}$), the Anomalous Distribution-aware Margin ($\{0.125, 0.25, 0.5\}$), and the minimum temperature of class-wise temperature ($\{0.8, 1.2\}$). For GraphSMOTE, we choose the GraphSMOTE\textsubscript{O} version which is without pretraining, as it shows excellent performance among multiple versions. For GraphENS, we set the feature masking hyperparameter $k$ at 0.01 and the temperature $\tau$ at 2. For GraphSHA, we keep the default setting in the code.

\textbf{Experimental Settings} To address the sparsity of the search space and the tendency for simulated formulas to exceed depth limits and include nonterminal nodes, we employ a larger number of simulations, specifically 100. We consider the maximum value simulated during these 100 simulations as the reward score for the state. In the context of citation network datasets, one trial comprises 100,000 episodes, and we conduct a single trial. For the Amazon co-purchase network, one trial involves 200,000 episodes, and we perform 10 trials. In the case of citation networks, all experiments are conducted ten times to calculate bAcc and F1 scores, while for the Amazon networks, each partition is executed three times. It's important to note that all experiments in this paper are performed using a single NVIDIA GeForce RTX 3090 GPU.

\textbf{GNN Settings} We perform experiments utilizing three prominent GNN models: GCN \cite{kipf2017semisupervised}, GAT \cite{veličković2018graph}, and SAGE \cite{hamilton2017inductive}. All three GNNs consist of a 2-layer neural network with a hidden layer dimension of 256. Other hyperparameters align with those detailed in \cite{song2022tam}. This includes employing the Adam \cite{kingma2014adam} optimizer, training the model for 2000 epochs, and selecting the optimal model parameters based on the average accuracy and F1 scores on the validation set. The initial learning rate is configured at 0.1 and undergoes halving when there is no improvement in the validation set loss for 100 consecutive generations. Additionally, weight decay is set to 0.0005 and is applied to all learnable parameters except for the last convolutional layer.

\begin{table}[!t]
\small
    \centering
    \caption{The searched loss functions, represented in the table, exemplify their effectiveness. For instance, the entry labeled ID A showcases the evaluation of node classification accuracy using GCN on the Cora training set, highlighting the high-performance loss function discovered by AutoLINC. Here, $\rho=10$.}
    \begin{tabular}{c|c|c|c}
    \toprule
ID	& Dataset & GNN	& Discovered Loss Function \\
\midrule
A &	Cora & GCN	& 
   $\exp(\tanh^{2}{\left( \frac{1}{N}\times \left(-y\right)+\hat{y}\right)})$ \\
B	 &CiteSeer & GCN&	$\left( -\left(N\times \hat{y} \right)+y\right)^{2}$\\
C	 &PubMed & GCN	&$\hat{y}+\left(\frac{1}{N}\times\left(-y\right)+\hat{y} \right)^{2}$\\
D	 &Cora & GAT	&$\left(-\tanh\left(\frac{1}{N}\times\left(-y\right)+\hat{y}\right)\right)^{2}$ \\
E	 &CiteSeer & GAT	&$\left(\tanh\left(-\hat{y}\right)+\tanh\left(y\right)\times\frac{1}{N} \right)^{2}$ \\
F	 &PubMed & GAT&	$\left(\exp(\frac{1}{N}\times\left(-y\right)\times2)+\hat{y}\right)^{2}$ \\
G	 &Cora & SAGE	&$\tanh^{2}\left(\tanh\left(\frac{1}{N}\times\left(-y\right)+\hat{y}\right)\right)$ \\
H	 &CiteSeer & SAGE	&$\left(\hat{y}+\left(-\frac{1}{\log\left(\tanh\left(y\right)\right)+N}\right)\right)^{2}$\\
I	 &PubMed & SAGE	&$\left(-\left(\tanh\left(y\right)+\hat{y}\times\left(-N\right)\right)\right)\times\hat{y}$\\
J	 &Amazon-computers&GCN	&$\left(\left(\hat{y}+N\right)+\log\left(\log\left(\frac{1}{\sqrt{y}}\right)\right)\right)^{2}$ \\
K	 &Amazon-photo&GCN	&$\left(\hat{y}+\sqrt{\sqrt{N\times log^{2}\left(y\right)}}\right)^{2}$\\
L	 &Amazon-computers&GAT	&$\log\left(\left(y+\left(-\hat{y}\right)\right)^{2}+\log\left(N\right)\right)$ \\
M	 &Amazon-photo&GAT	&$\left|-y+\left(\hat{y}+\tanh^{2}\left(N\right)\right) \right|$ \\
N	 &Amazon-computers&SAGE	&$\left(\hat{y}+\left(N+\log\left(\log\left(\frac{1}{\sqrt{y}}\right)\right)\right)\right)^{2}$ \\
O	 &Amazon-photo&SAGE	&$\hat{y}\times\left(\hat{y}+\left(N+\log\left(-\log\left(y\right)\right)\right)\right)$\\
\bottomrule
\end{tabular}
\vskip -0.2in
\label{tab:exp}
\end{table}

\begin{table}[!t]
\small
    \centering
    \caption{The searched loss functions by AutoLINC combined with GraphSHA. The evaluation results of these loss functions are presented in Table \ref{tab:6}. }
    \begin{tabular}{c|c|c|c}
    \toprule
Frame	& Dataset & GNN	& Discovered Loss Function \\
\midrule
AUTOLINC(GraphSHA)   &	Cora & GCN	& $\hat{y} \times ((\hat{y}-y)\times N + \hat{y})$\\
AUTOLINC(GraphSHA)   & CiteSeer & GCN& $(y-N \times \hat{y}^3)^2$ \\ 
AUTOLINC(GraphSHA)	 &PubMed & GCN	&  $\hat{y} + \frac{1}{\exp(N \times \hat{y} \times y)}$\\ 
AUTOLINC(GraphSHA)	 &Cora & GAT	& $N+(y+\sqrt{-\hat{y}})^2$\\
AUTOLINC(GraphSHA)	 &CiteSeer & GAT	& $N\times (y-\hat{y})^4$ \\
AUTOLINC(GraphSHA)	 &PubMed & GAT&	 $ \hat{y} + y \times \exp(N \times |\hat{y}|^2)$\\
AUTOLINC(GraphSHA)	 &Cora & SAGE	& $N \times |\hat{y}-\tanh(\hat{y}+y)|$\\
AUTOLINC(GraphSHA)	 &CiteSeer & SAGE & $N \times \tanh^2(\tanh(y^2 - \hat{y}))$\\
AUTOLINC(GraphSHA)	 &PubMed & SAGE	& $\frac{|\hat{y}|}{N\times \sqrt{\hat{y}+y}}$\\
\bottomrule
\end{tabular}
\vskip -0.2in
\label{tab:exp3}
\end{table}

\begin{table}[!t]
\small
    \centering
    \caption{The searched loss functions by AutolossZero and AutoLINC. The evaluation results of these loss functions are presented in Table \ref{non-loss}. }
    \begin{tabular}{c|c|c|c}
    \toprule
Frame	& Dataset & GNN	& Discovered Loss Function \\
\midrule
AUTOLOSS-ZERO-A &	Cora & GCN	& $\left(\left( \hat{y}+\log{y} \right) \times \left( \tanh{1} + \hat{y} + y\right)\right)^2$ \\
AUTOLOSS-ZERO-A & CiteSeer & GCN&	$\tanh{\left(\tanh{\left( -\hat{y} \right)}\right)}+\tanh{\sqrt{\hat{y}+y}}$\\
AUTOLOSS-ZERO-A	 &PubMed & GCN	&$\left|1+y\right| \times \left( \hat{y}-y\right) \times \hat{y}^8$\\
AUTOLOSS-ZERO-A	 &Cora & GAT	&$\tanh{\left( \sqrt{\hat{y}}+y \right)}+\left( \left|\hat{y} \right| \times \exp(y)\right)^2 $ \\
AUTOLOSS-ZERO-A	 &CiteSeer & GAT	& $\frac{\left( \left(y^2-y \right)\times \left( -y\times \sqrt{y}\right)\times \log(\log(1)) \right)}{\exp}$\\
AUTOLOSS-ZERO-A	 &PubMed & GAT& $\left(\left(y+1 \right)\times \hat{y} \right)^2 + \left( \exp+\sqrt{y}\right)$\\
AUTOLOSS-ZERO-A	 &Cora & SAGE	& $\left(\left(\tanh{\hat{y}}+\hat{y}^2 \right)-2\times y \right)^2$ \\
AUTOLOSS-ZERO-A	 &CiteSeer & SAGE	& $\left(y+\sqrt{\hat{y}+y}+\exp{\sqrt{-\hat{y}}} \right)$\\
AUTOLOSS-ZERO-A	 &PubMed & SAGE	& $\left|y\times \hat{y}-2\right|$\\
AUTOLOSS-ZERO-B &	Cora & GCN	& $\left( \exp(y) \times \hat{y}^2 +\tanh{\hat{y}}\right) \times \left( y \times \left| N \right| + y + N + \hat{y}\right)$\\
AUTOLOSS-ZERO-B & CiteSeer & GCN& $\tanh((-y+\hat{y}) \times (N+2 \times \hat{y}))$\\
AUTOLOSS-ZERO-B	 &PubMed & GCN	& $ -\sqrt{y} \times (\tanh{\hat{y}} + \hat{y}) + (\log{N}+N) \times \hat{y}^2$\\
AUTOLOSS-ZERO-B	 &Cora & GAT	& $ \left( \tanh(\exp(\hat{y})) + 2\times y \times \hat{y} \right) \times \left(N+y+\hat{y} \right) \times  \hat{y}$\\
AUTOLOSS-ZERO-B	 &CiteSeer & GAT	& $\sqrt{\hat{y}+N} \times \tanh{\hat{y}} + \hat{y}^2 \times y \times \sqrt{N}$ \\
AUTOLOSS-ZERO-B	 &PubMed & GAT& $\frac{1}{N \times \tanh{\hat{y}} + \hat{y} + \frac{1}{y}}$\\
AUTOLOSS-ZERO-B	 &Cora & SAGE	& $2\times \hat{y} \times (\tanh(N \times \hat{y}) - y + \tanh{\hat{y}}) $ \\
AUTOLOSS-ZERO-B	 &CiteSeer & SAGE	& $|\hat{y}|+\hat{y}+(y+\hat{y}) \times y \times \hat{y} + 2\times N \times \sqrt{\exp(\hat{y})}$\\
AUTOLOSS-ZERO-B	 &PubMed & SAGE	& $\hat{y}^4 \times \tanh{\hat{y}} \times N - \tanh(y \times \hat{y})$\\
AUTOLINC(1 HOUR) &	Cora & GCN	& $(\tanh(N \times \hat{y}) - y)^2$\\
AUTOLINC(1 HOUR) & CiteSeer & GCN& $(y-N \times \hat{y})^2$ \\ 
AUTOLINC(1 HOUR)	 &PubMed & GCN	&  $N \times \hat{y}^2 - y \times \hat{y}$\\ 
AUTOLINC(1 HOUR)	 &Cora & GAT	& $\sqrt{\hat{y}} \times (\hat{y}-\frac{y}{N})$\\
AUTOLINC(1 HOUR)	 &CiteSeer & GAT	& $\hat{y}^2-\frac{y \times \hat{y}}{N}$ \\
AUTOLINC(1 HOUR)	 &PubMed & GAT&	 $\exp(-\hat{y} \times (y+N)) \times y$\\
AUTOLINC(1 HOUR)	 &Cora & SAGE	& $(\tanh(N \times \hat{y}) + \sqrt{-y})^2$\\
AUTOLINC(1 HOUR)	 &CiteSeer & SAGE & $N \times \hat{y}^2 - y \times \hat{y}$\\
AUTOLINC(1 HOUR)	 &PubMed & SAGE	& $N \times \hat{y}^2 - y \times \hat{y}$\\
\bottomrule
\end{tabular}
\vskip -0.2in
\label{tab:exp2}
\end{table}

\begin{figure}[htb]
\centering
\vskip -0.1in
\subfloat[A] {\includegraphics[width=0.33\linewidth]{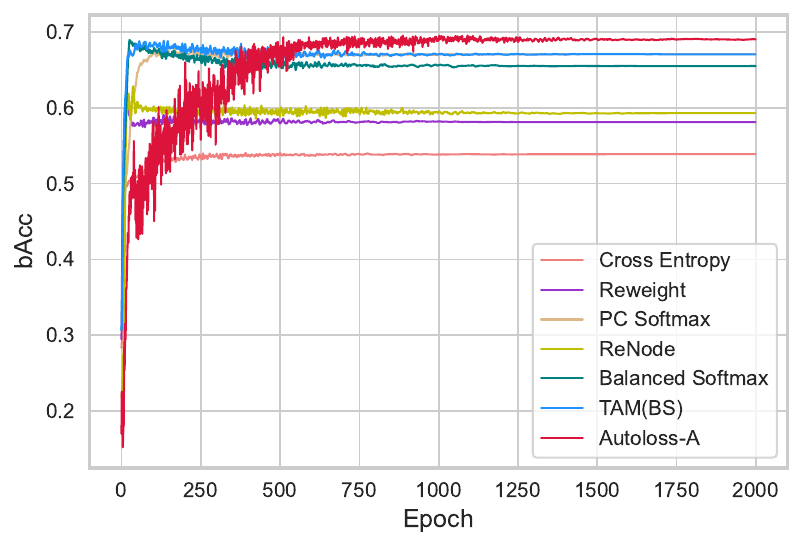}}
\subfloat[B] {\includegraphics[width=0.33\linewidth]{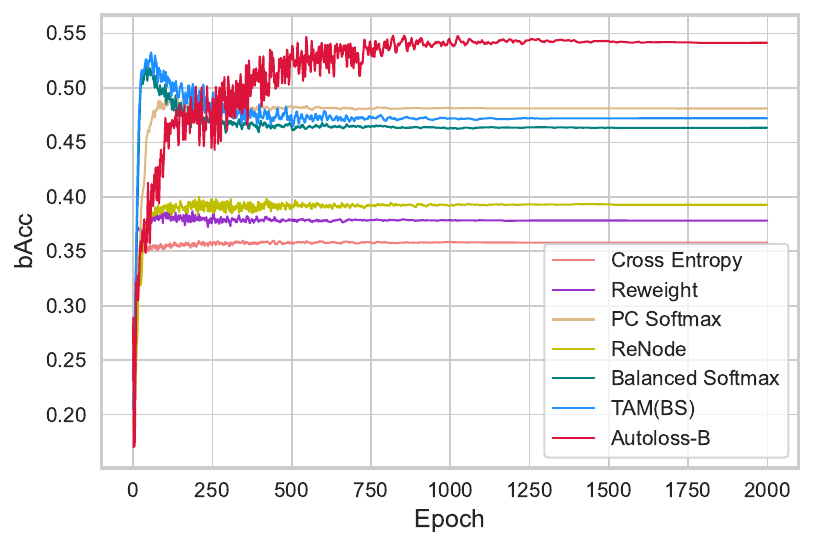}}
\subfloat[C] {\includegraphics[width=0.33\linewidth]{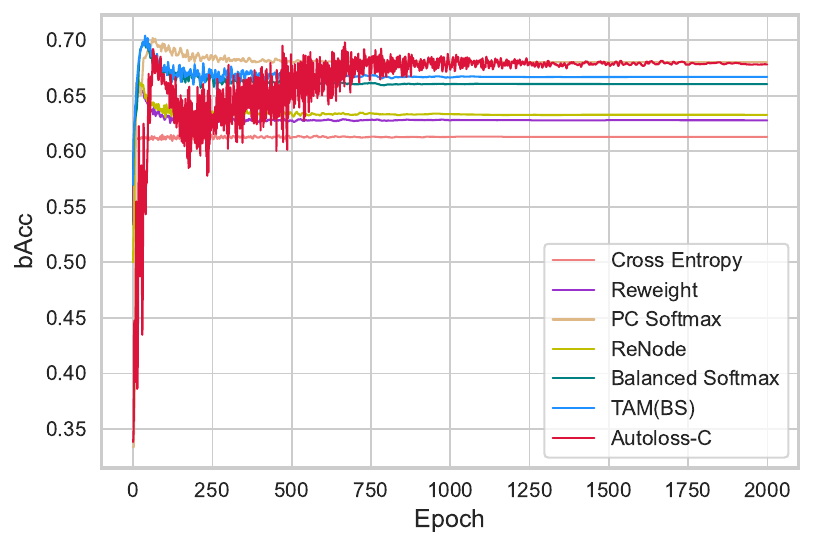}}\\
\vskip -0.1in
\subfloat[D] {\includegraphics[width=0.33\linewidth]{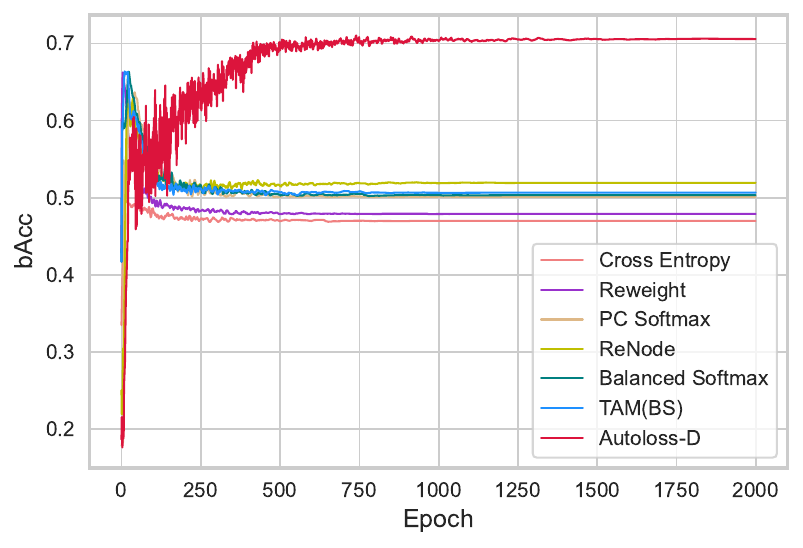}}
\subfloat[E] {\includegraphics[width=0.33\linewidth]{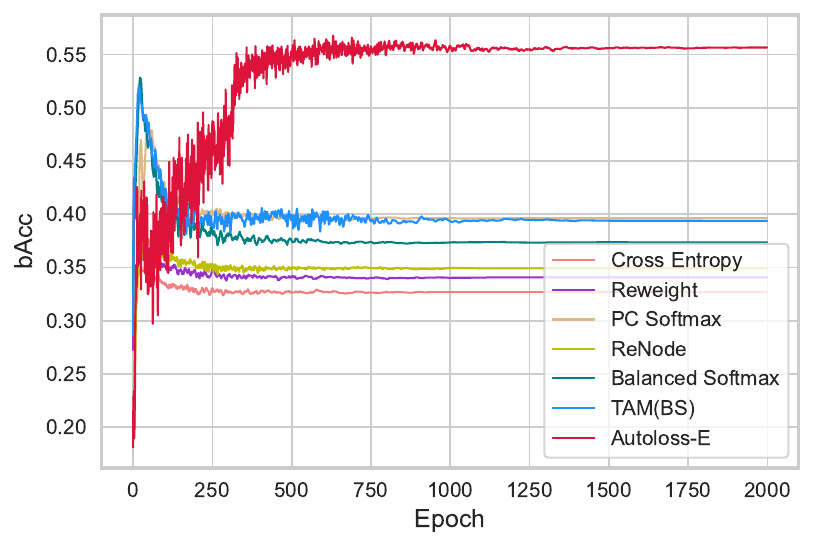}}
\subfloat[F] {\includegraphics[width=0.33\linewidth]{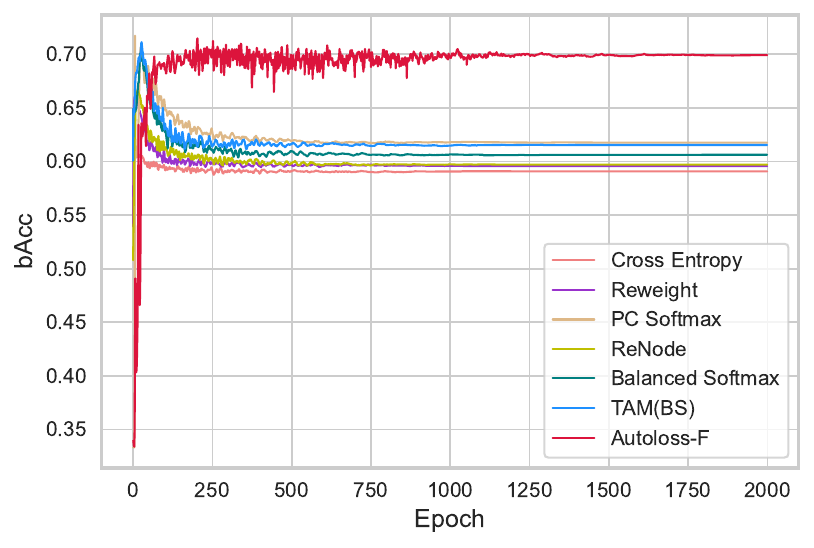}}\\
\vskip -0.1in
\subfloat[G] {\includegraphics[width=0.33\linewidth]{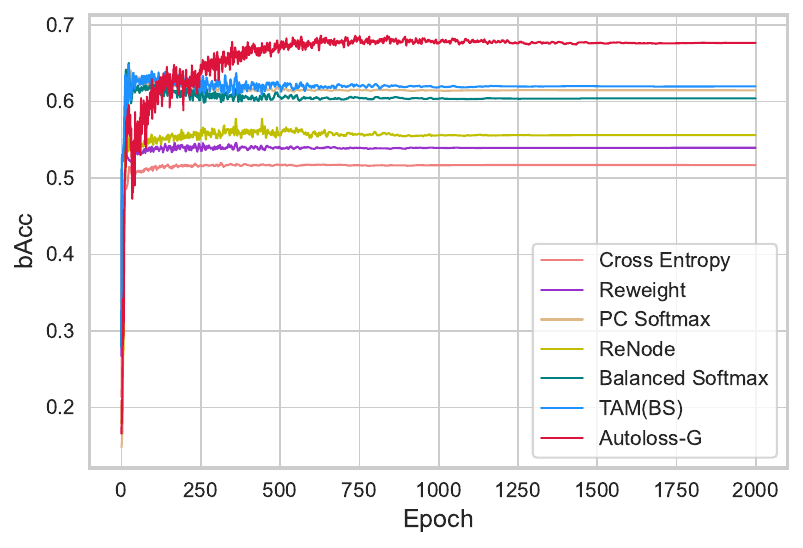}}
\subfloat[H] {\includegraphics[width=0.33\linewidth]{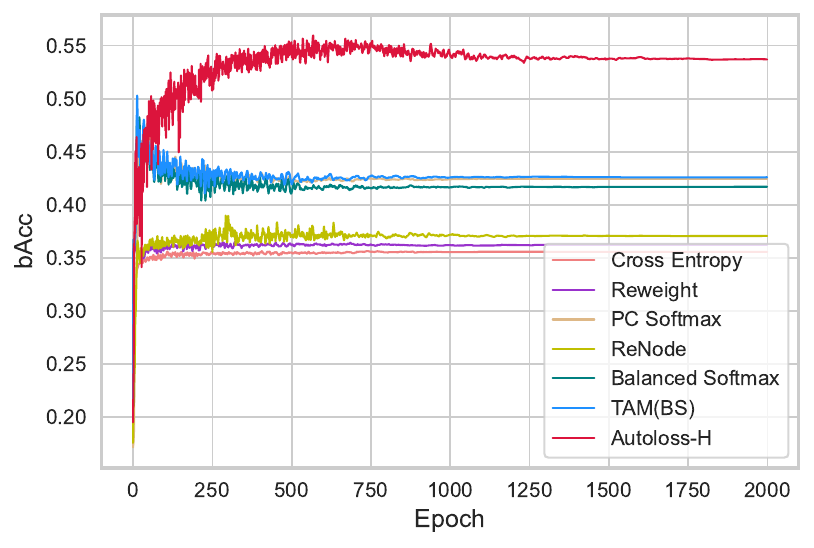}}
\subfloat[I] {\includegraphics[width=0.33\linewidth]{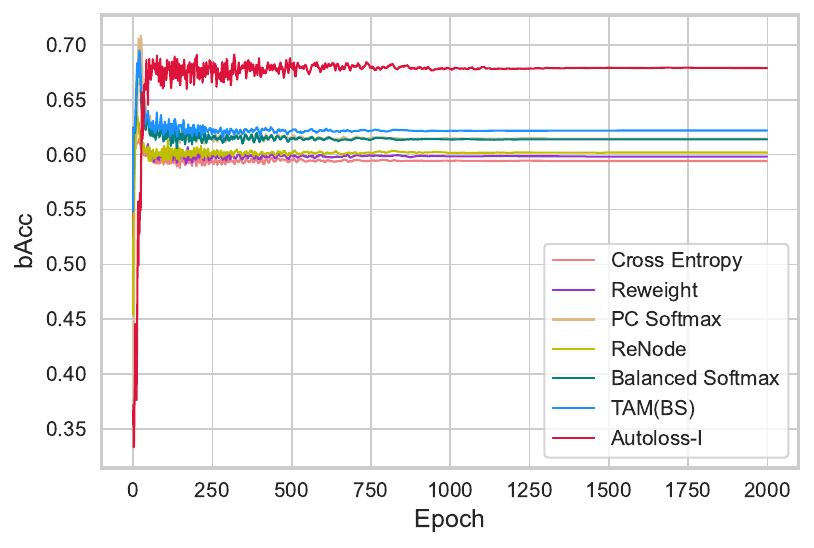}}\\
\vskip -0.1in
\subfloat[J] {\includegraphics[width=0.33\linewidth]{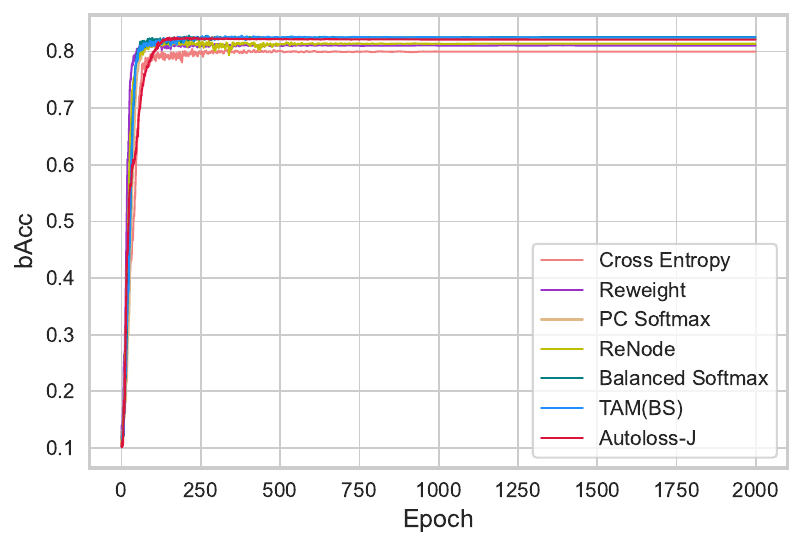}}
\subfloat[K] {\includegraphics[width=0.33\linewidth]{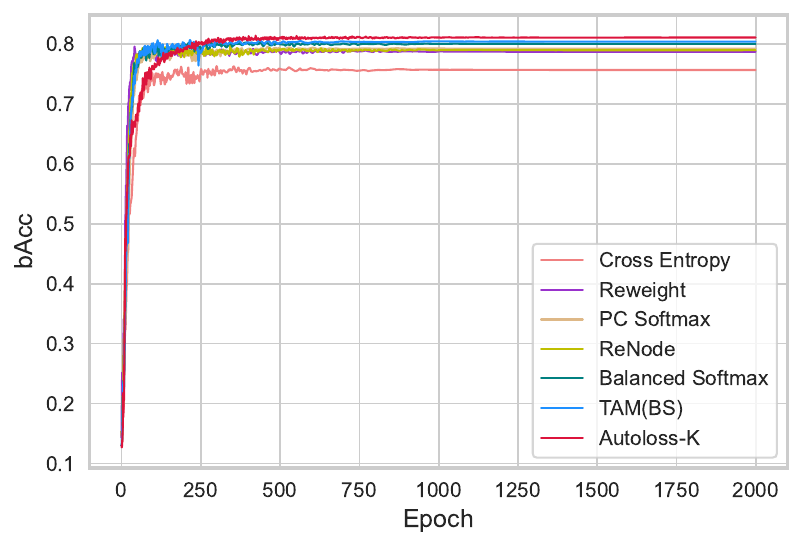}}
\subfloat[L] {\includegraphics[width=0.33\linewidth]{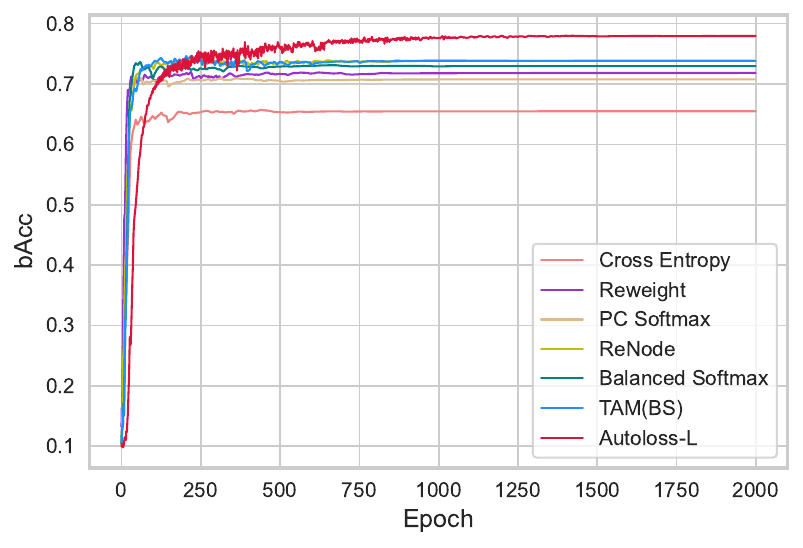}}\\
\vskip -0.1in
\subfloat[M] {\includegraphics[width=0.33\linewidth]{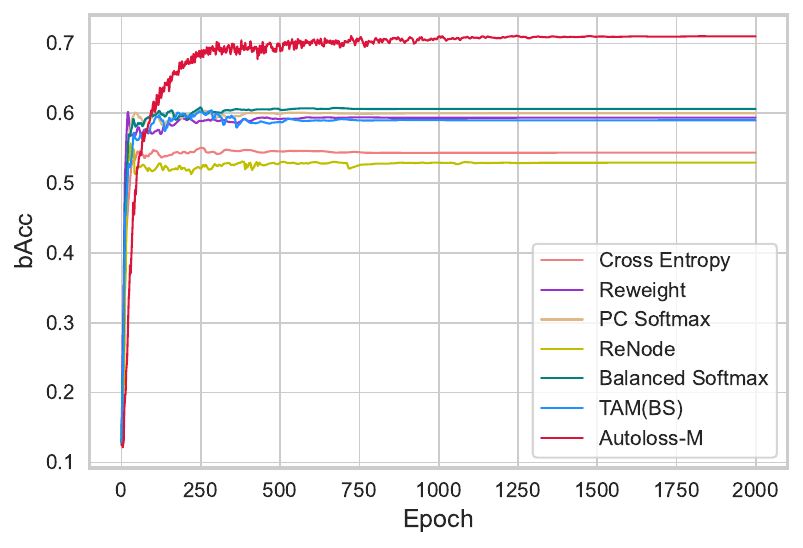}}
\subfloat[N] {\includegraphics[width=0.33\linewidth]{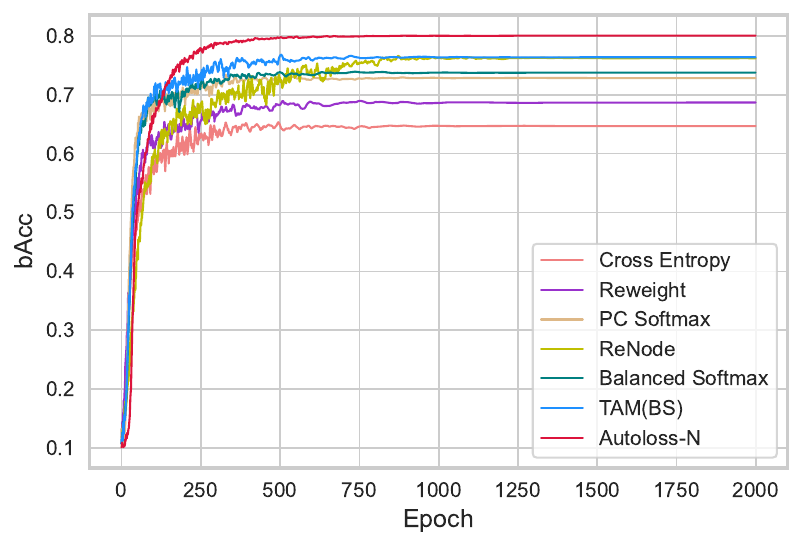}}
\subfloat[O] {\includegraphics[width=0.33\linewidth]{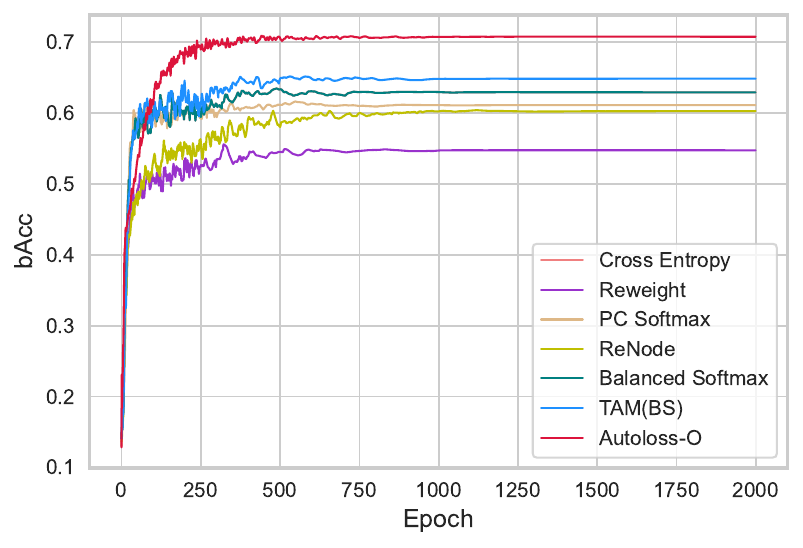}}\\
\vskip -0.1in
\caption{The convergence of loss function discovered in Table \ref{tab:exp}.}
\label{fig:lc}
\end{figure}

\begin{figure}[htb]
\vskip -0.1in
\subfloat[Cora] {\includegraphics[width=0.33\linewidth]{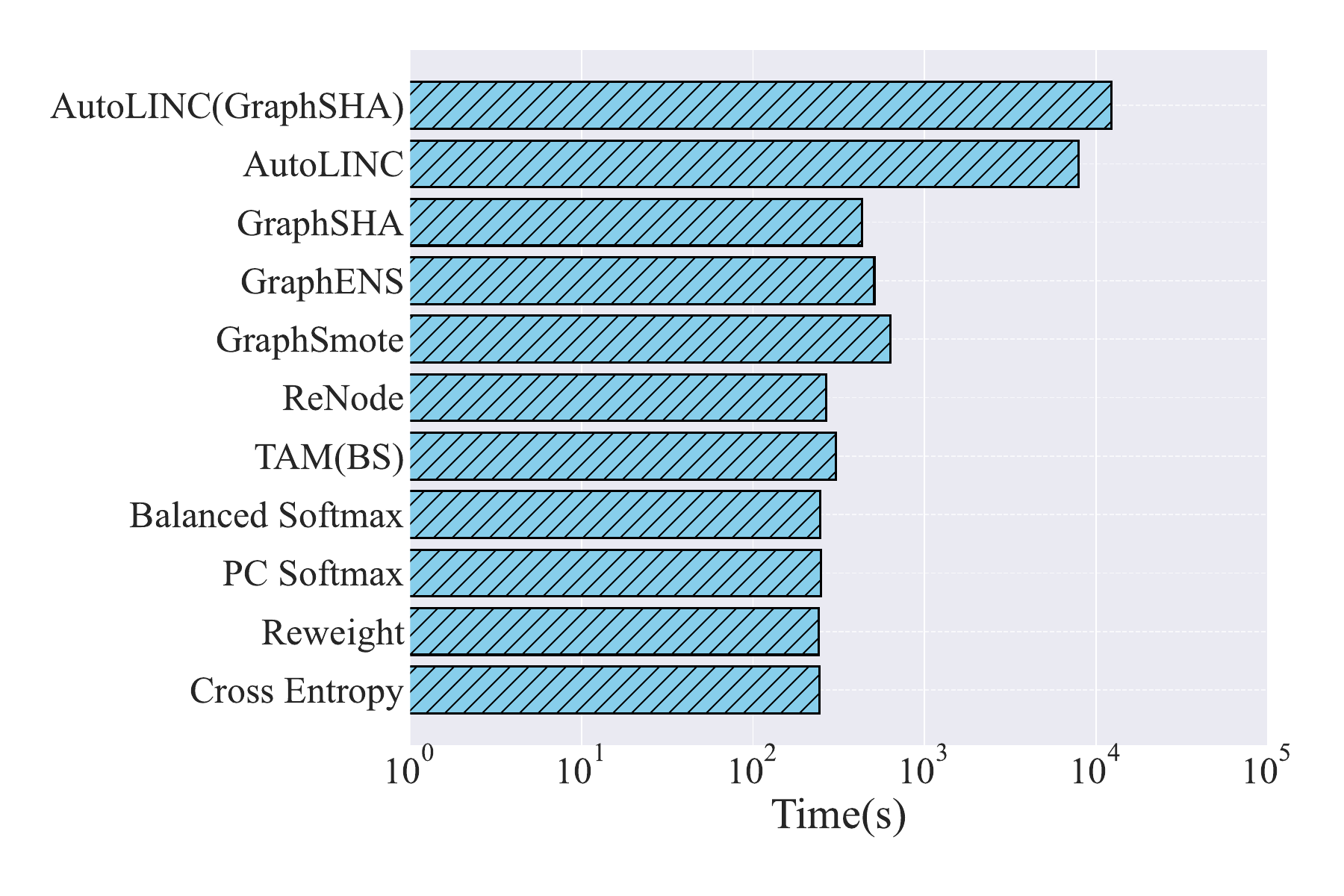}}
\subfloat[CiteSeer] {\includegraphics[width=0.33\linewidth]{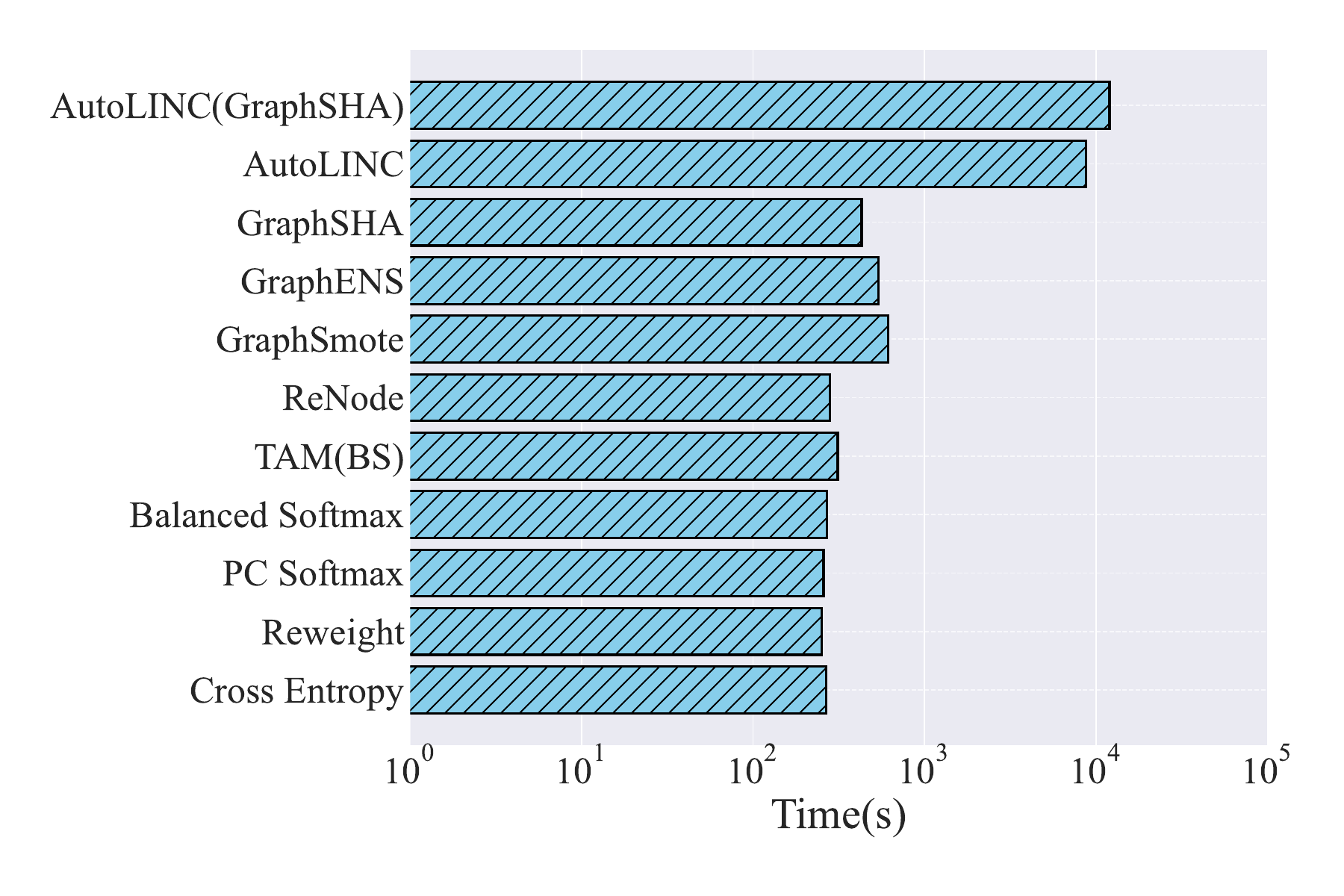}}
\subfloat[PubMed] {\includegraphics[width=0.33\linewidth]{PubMed_GCN.pdf}}\\
\vskip -0.1in
\caption{The runtime on GCN across three citation network datasets using 10 random seeds.}
\label{fig:time_all}
\end{figure}

\end{document}